\documentclass[10pt,twocolumn,letterpaper]{article}

\usepackage{multirow}
\usepackage[pagenumbers]{wacv} 

\usepackage{algorithm}
\usepackage{algorithmic}
\usepackage{amsmath}

\definecolor{wacvblue}{rgb}{0.21,0.49,0.74}
\usepackage[pagebackref,breaklinks,colorlinks,allcolors=wacvblue]{hyperref}

\def\wacvPaperID{1969} 
\def\confName{WACV}
\def\confYear{2027}

\title{SHIFT: Steering Hidden Intermediates in Flow Transformers}

\author{Nina Konovalova\\
MBZUAI, FusionBrain Lab\\
\and
Ibragim Idrisov\\
Lomonosov Moscow State University\\
\and
Aibek Alanov\\
HSE, FusionBrain Lab\\
}

\begin{document}
\maketitle
\begin{abstract}
Diffusion models have become the leading approach for high-fidelity image generation, with recent DiT-based models achieving strong prompt adherence and high visual quality. However, removing unwanted concepts from these models typically requires expensive gradient-based fine-tuning, which is particularly prohibitive for multi-billion-parameter Diffusion Transformers and incompatible with distilled variants. We propose SHIFT, a simple model-weight-update-free inference-time framework for concept suppression in DiT models via targeted manipulation of intermediate activations at inference time, inspired by activation steering in LLMs. SHIFT learns steering vectors from contrastive prompts and applies them dynamically. We demonstrate that a single time-independent steering vector generalizes across all denoising timesteps and that vectors computed on a distilled model transfer effectively to its non-distilled counterpart without recomputation. SHIFT achieves strong concept suppression while preserving prompt fidelity and image quality. The same mechanism also enables flexible control — adding, removing, or changing objects and styles without retraining. SHIFT targets hosted inference settings: it does not remove the concept from the model weights and is therefore complementary to weight-level unlearning. The project page is available at \url{https://controlgenai.github.io/SHIFT}.
\end{abstract}
    
\section{Introduction}
\label{sec:intro}
Diffusion models~\cite{ho2020denoising,rombach2022high} have established a new state-of-the-art in high-fidelity text-to-image synthesis~\cite{podell2023sdxl}. As their capabilities scale, so do the risks associated with generating harmful, copyrighted, or otherwise prohibited content. This has motivated the development of Concept Erasure (CE): the task of reliably suppressing specific semantic concepts within a model's generative manifold while preserving everything else.

Most existing erasure strategies for text-to-image diffusion models rely on model optimization. Approaches such as ESD~\cite{gandikota2023erasing}, Concept Ablation (CA)~\cite{kumari2023ablating}, and Erase-and-Preserve (EAP)~\cite{bui2024erasing} require gradient-based tuning to achieve effective concept removal. While practical for earlier UNet architectures (typically under 1B parameters), these methods become prohibitively expensive on modern Diffusion Transformers (DiTs) such as Flux~\cite{labs2025flux}, which contains 12B parameters. Moreover, methods like ESD or EAP frequently utilize negative guidance during the optimization phase to steer the model distribution away from the target concept. This dependency presents a significant challenge for distilled models such as Flux.1[schnell], because these models are guidance-distilled to operate at a fixed guidance scale. Non-optimization alternatives, such as Unified Concept Editing (UCE)~\cite{gandikota2024unified}, propose closed-form weight edits that avoid retraining. However, these edits are often insufficient for concrete object erasure in unified architectures.

In parallel, activation steering has emerged as a lightweight yet potent control mechanism for Large Language Models (LLMs)~\cite{liu2025reducing,turner2023steering,beaglehole2026toward}. Representation Engineering~\cite{zou2023representation} and related work show that high-level semantic behavior can be modulated by injecting linear directions into the latent representation at inference time, with no weight modification. 
Recent works have begun extending this paradigm to diffusion models~\cite{gaintseva2025casteer}, showing promising results for concept control via contrastive activation differences. This raises a natural question: can effective, model-weight-update-free concept removal be achieved in large-scale Diffusion Transformers by directly steering their unified internal activations?

We introduce \text{SHIFT}, a simple yet effective steering framework specifically engineered for MM-DiT architectures. SHIFT learns steering vectors from contrastive prompt pairs and applies them dynamically during inference to suppress unwanted concepts while preserving prompt fidelity and image quality. Beyond suppression, the same mechanism enables flexible concept manipulation -- adding, removing, or switching objects and visual styles -- all without updating the diffusion-model weights. On the I2P benchmark with Flux.1[schnell], SHIFT reduces the number of NudeNet detections from 612 to 105 (a $5.8\times$ reduction), resulting in more than $3\times$ fewer residual detections compared to the best fine-tuning baseline. At the same time, CLIP and FID scores on MS-COCO remain nearly unchanged.


Our contributions are as follows:

\begin{itemize}

    \item \textbf{Unified steering framework.} SHIFT introduces a systematic activation-steering framework that targets the post-block text stream of MM-DiT models and combines token-wise directions with concept-specific prompt gating. It controls both non-distilled FLUX.1[dev] and distilled FLUX.1[schnell] at inference time without updating their weights.
    \item \textbf{Temporal invariance in MM-DiT representation space.} We demonstrate that a single steering vector computed at the first denoising step performs on par with per-step adaptation, revealing remarkable stability of semantic structure throughout the diffusion process.
     \item \textbf{Cross-distillation transfer.} Steering vectors derived from the distilled model transfer effectively to its non-distilled counterpart, enabling zero-shot transfer to FLUX.1[dev] without using any FLUX.1[dev] activations during direction extraction.
    \item \textbf{Intervention location.}  Based on analysis of the residual structure within double-stream blocks, we find that the post-block text state serves as the effective steering location. This was confirmed empirically by comparing it with alternative sites.
    \item \textbf{Extensive analysis.} We provide a detailed study of spatial and temporal dynamics of steering in DiTs.
    
\end{itemize}


\section{Related Work}
\subsection{Concept erasure and safety}
Recent state-of-the-art text-to-image diffusion models trained on large, imperfectly filtered datasets such as LAION~\cite{schuhmann2022laion} can generate inappropriate or copyrighted content. One mitigation is dataset filtering~\cite{rombach2022high} or post-generation filtering~\cite{rando2022red}. However, these approaches do not fully prevent harmful content and require additional processing. Many works therefore focus on concept erasure (CE) while preserving overall generation quality. Methods such as ESD~\cite{gandikota2023erasing} employ fine-tuning to unlearn concepts using negative guidance from a frozen teacher model, while Concept Ablation (CA)~\cite{kumari2023ablating} optimizes cross-attention layers to redirect target concepts toward neutral anchors. Subsequent approaches like Unified Concept Editing (UCE)~\cite{gandikota2024unified} introduce closed-form edits to attention projections for efficient, training-free erasure, though they may struggle to fully remove concrete objects. For modern architectures, EraseAnything~\cite{gao2025eraseanything} adapts erasure to rectified flow models via LoRA tuning and attention regularization, and Erasing with Adversarial Preservation (EAP)~\cite{bui2024erasing} incorporates adversarial concept identification to preserve unrelated generations during fine-tuning. However, these methods have limitations on modern diffusion models, including optimization overhead, reliance on guidance that must be adapted for distilled models, or dependence on LLM agents.

\subsection{Activation steering in Large Language Models.}
Steering in large language models (LLMs) has emerged as a practical paradigm for controlling behavior. Various approaches have been explored, including specific interventions on weights~\cite{ziegler2019fine,ilharco2022editing,meng2022locating}, prompt engineering~\cite{zhou2022steering}, and soft prompting~\cite{khashabi2022prompt,lester2021power}. At the same time, one of the most promising directions involves modifying activations using specifically calculated steering vectors. These vectors can be estimated using gradient descent~\cite{hernandez2023inspecting,subramani2022extracting}, PCA decomposition~\cite{zou2023representation}, or simply as the mean difference between activations from contrastive prompt pairs~\cite{turner2023steering,li2023inference}. While these methods operate primarily on transformer-based LLMs, they provide foundational insights for extending activation steering to other architectures, such as diffusion models.

\subsection{Latent space navigation and Diffusion Steering.}


Diffusion models support controllable generation through various forms of guidance and latent-space manipulation, including classifier guidance~\cite{dhariwal2021diffusion}, classifier-free guidance~\cite{ho2022classifier}, and semantic guidance~\cite{brack2022stable}. More recently, several works have adapted activation-level interventions directly inside diffusion models. In particular, methods operating on U-Net architectures have shown success in manipulating cross-attention maps using contrastive prompts for tasks such as concept erasure and style transfer~\cite{gaintseva2025casteer,rodriguez2024controlling}.

Our work focuses on modern MM-DiT architectures such as Flux.
The closest prior approach is CASteer~\cite{gaintseva2025casteer},
which injects contrastive directions into cross-attention outputs used in U-Net or a CA-based DiT. MM-DiT exposes no such module, so the intervention site has to be reconsidered rather than transferred. In contrast to prior activation-based approaches that primarily target U-Nets or rely on per-step adaptation, SHIFT introduces a unified, model-weight-update-free steering framework that exploits the specific residual structure of double-stream transformer blocks, keeps a token-wise field over the text stream rather than a single pooled direction, and modulates it by prompt relevance so that several concepts can be controlled independently. We further demonstrate distinctive properties of DiT activation
spaces, including temporal invariance and cross-distillation transfer.


\section{Preliminaries}
\label{sec:prelim}

Early text-to-image diffusion models were mostly U-Net based, with explicit self-attention over image latents and cross-attention for text conditioning. Modern text-to-image systems increasingly use diffusion transformers (DiTs), where image latents and text conditioning are represented as token sequences and processed by transformer blocks. In this work, we focus on Flux.1. At each denoising step, the Flux transformer receives image latent tokens together with two text-conditioning inputs: token-level text embeddings, and a pooled text embedding. In the standard Flux.1 pipeline, the token-level embeddings are produced by T5 \cite{raffel2020exploring}, while the pooled embedding is produced by the CLIP text branch \cite{radford2021learning}. The pooled embedding is combined with the diffusion timestep and used to modulate the transformer computation.

Let $X_i^\ell$ and $X_t^\ell$ denote the image and text streams entering block $\ell$, and let $p$ denote the pooled text embedding. The pooled route is first combined with the timestep $\tau$ into a modulation vector

\begin{equation}
    e_\tau = \phi(\tau, p),
    \label{eq:flux_timestep_pooled}
\end{equation}

which parametrizes adaptive normalization and gating inside transformer blocks. Thus, Flux uses token-level text embeddings for sequence-level prompt semantics and a pooled text route for global conditioning.

Flux contains two types of transformer blocks: \textbf{double-stream} and \textbf{single-stream}. In double-stream blocks, text and image tokens are kept as separate residual streams that interact through joint attention. A double-stream block first applies a gated joint-attention residual update and then applies a modality-specific feed-forward residual update. These internal operations could be abstracted as a total residual update for each stream:

\begin{equation}
\begin{aligned}
    X_i^{\ell+1} &= X_i^\ell + R_{i,\ell}(X_i^\ell, X_t^\ell; e_\tau), \\
    X_t^{\ell+1} &= X_t^\ell + R_{t,\ell}(X_t^\ell, X_i^\ell; e_\tau),
\end{aligned}
\label{eq:double_stream_residual}
\end{equation}

where $R_{i,\ell}$ and $R_{t,\ell}$ denote the total residual updates produced by the block for the image and text streams, respectively. Each residual update contains the block's joint-attention and feed-forward computations, including the adaptive modulation determined by $e_\tau$. The outputs $X_t^{\ell+1}$ and $X_i^{\ell+1}$ are the post-block states propagated to subsequent blocks. An illustration of the block structure is provided in Appendix.

By contrast, single-stream blocks concatenate text and image tokens and process
them with shared weights. In this work, we primarily steer double-stream blocks,
where text and image streams remain explicitly separated.

\section{Method}
\label{sec:method}

\begin{figure*}[t]
    \centering
    \includegraphics[width=\linewidth]{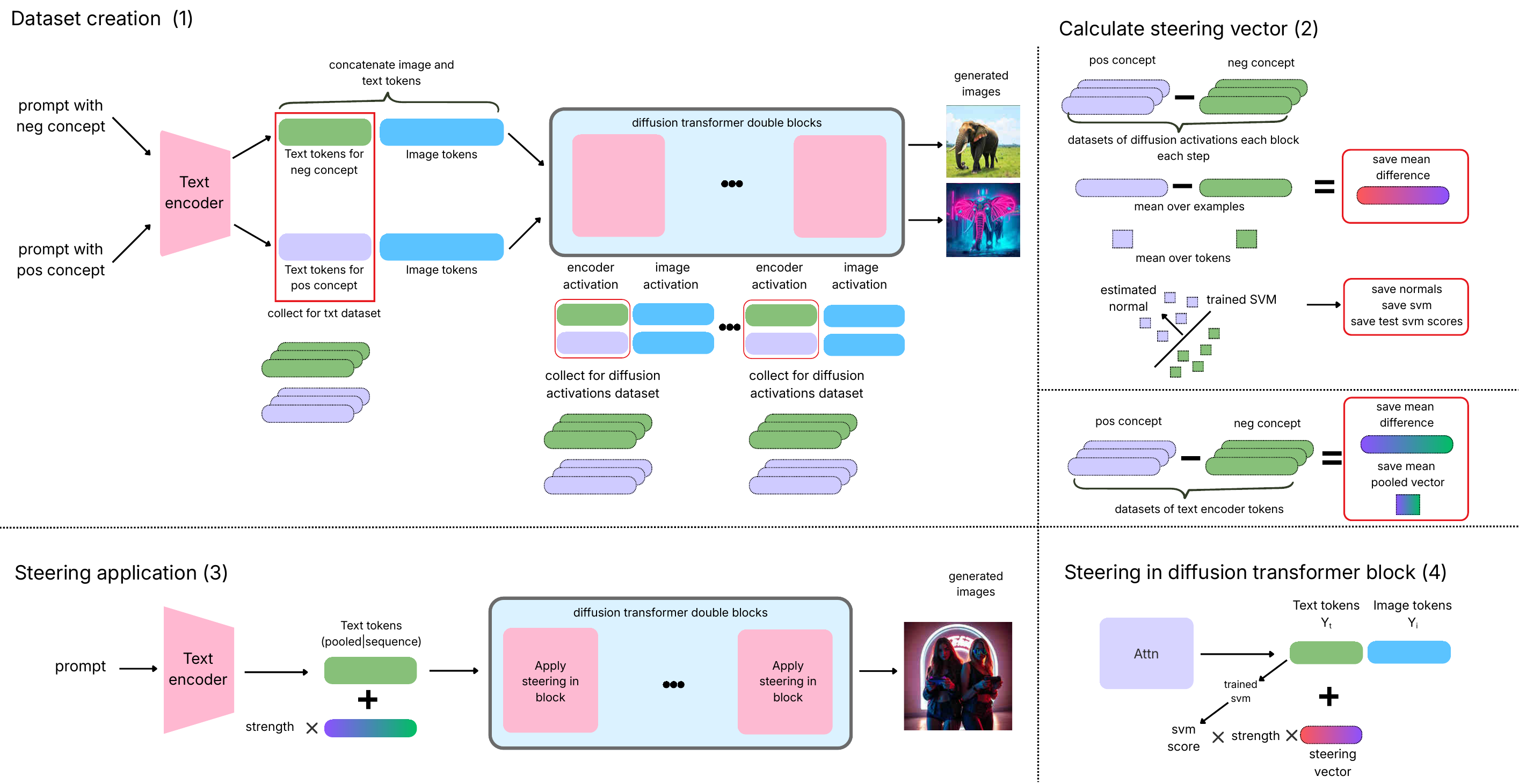}
    \caption{Overview of the steering pipeline: (1) dataset construction from contrastive prompt pairs, (2) steering vector computation based on mean difference and separation plane, (3) application of the vector during inference, and (4) steering inside the diffusion transformer.}
    \label{fig:main_scheme}
    \vspace{-8pt}
\end{figure*}

In this work, we introduce SHIFT, a model-weight-update-free activation steering framework for modern Diffusion Transformers. We primarily target concept erasure in Flux.1 models (both the distilled Flux.1[schnell] and the full Flux.1[dev]), while also demonstrating the generality of our approach to concept addition, removal, and style manipulation. The method consists of three main stages: (1) identifying the optimal intervention site within the DiT architecture, (2) constructing concept-specific steering vectors from contrastive activations, and (3) applying these vectors at inference time with adaptive gating. An overview of the full pipeline is shown in Figure~\ref{fig:main_scheme}.


\subsection{Where to apply steering}

The residual structure of Flux’s double-stream blocks (Eq.~\ref{eq:double_stream_residual}) offers multiple candidate intervention points. For the text stream, one can steer either the internal residual update $R_{t,\ell}$ or the post-block output state $X_t^{\ell+1}$.
These choices are not equivalent under difference-based steering. Consider a single contrastive pair, with $X_t^{-}$ and $X_t^{+}$ denoting the source and target text states entering the block, and $R_t^{-}$ and $R_t^{+}$ the corresponding residual updates. Steering only the residual branch yields:
\begin{equation}
    \widetilde R_t = R_t^{-} + (R_t^{+}-R_t^{-}) = R_t^{+}.
\end{equation}
resulting in the block output
\begin{equation}
    \widetilde X_t^{\ell+1} = X_t^{-}+\widetilde R_t = X_t^{-}+R_t^{+}.
\end{equation}

However, the desired target state is

\begin{equation}
    X_t^{\ell+1,+} = X_t^{+}+R_t^{+}.
\end{equation}

Thus, even exact residual-branch steering leaves the mismatch

\begin{equation}
    X_t^{\ell+1,+} - \widetilde X_t^{\ell+1} = X_t^{+}-X_t^{-}.
    \label{eq:residual_branch_mismatch}
\end{equation}










In contrast, steering the post-block state directly resolves this issue. In the single-pair case,

\begin{equation}
    \widetilde X_t^{\ell+1} = X_t^{\ell+1,-} + \left( X_t^{\ell+1,+} - X_t^{\ell+1,-} \right) = X_t^{\ell+1,+}.
    \label{eq:post_block_exact}
\end{equation}

Thus, post-block steering exactly reconstructs the target activation for each contrastive pair. When using a mean direction across multiple pairs, it performs approximate distribution transport toward the desired concept-free (or concept-modified) region. Beyond mathematical exactness, post-block steering is more architecture-agnostic, as block outputs serve as a standard interface across residual transformers, whereas internal attention and feed-forward implementations vary. We therefore adopt post-block text states $X_t^{\ell+1}$ in selected double-stream blocks as our primary steering site. We validate this choice empirically in Sec.~\ref{sec:where}: compared to intervening on the joint-attention output $Y_t$ (Eq.~\ref{eq:Steered}), post-block steering reaches comparable or stronger concept suppression and incurs less degradation of non-target concepts (Tables~\ref{tab:ablation_place_nudity},~\ref{tab:ablation_place_snoopy}).

As a complementary pathway, we also steer the pooled text embedding that forms the global modulation vector $  e_\tau  $. This dual intervention allows us to control both fine-grained token semantics and global conditioning.



\subsection{Steering vector construction}
We construct a separate steering vector for each target concept $X$ and for each steering location (text encoder pooled embedding and diffusion transformer tokens). We collect a small paired dataset of $n$ prompt pairs $(p_k^{-}, p_k^{+})$, where $p_k^{+}$ adds the target concept to an otherwise neutral prompt, ``an elephant'' vs.\ ``an elephant in cyberpunk style''. Running the model on these prompts, we record activations at the chosen locations and obtain paired samples $(X_k^{-}, X_k^{+})$. The process of dataset construction is illustrated in Fig.~\ref{fig:main_scheme} (1). \\
After dataset collection, we estimate the steering vectors (Fig.~\ref{fig:main_scheme} (2)).

\textbf{Text encoder steering vector.}
For the pooled text-encoder representation, we use the raw activation-difference vector as the steering direction.

\textbf{Diffusion transformer steering vector.}
For diffusion-transformer steering, we use the mean-difference direction at the
post-block text states defined in Sec.~\ref{sec:prelim}. For prompt pair $k$,
denote the recorded post-block text activations at block $\ell$ and denoising
step $\tau$ by

\begin{equation}
    X_{k,\ell,\tau}^{-}, X_{k,\ell,\tau}^{+} \in \mathbb{R}^{T\times C}.
\end{equation}

Here $T$ is the number of text tokens and $C$ is the channel dimension. We
estimate a per-token steering field

\begin{equation}
    \Delta X_{\ell,\tau} = \frac{1}{n} \sum_{k=1}^{n} \left( X_{k,\ell,\tau}^{+} - X_{k,\ell,\tau}^{-} \right) \in \mathbb{R}^{T\times C}.
    \label{eq:hidden_mean_diff}
\end{equation}


To separate direction from strength, we normalize the field over the channel
dimension for each token:

\begin{equation}
    v_{\ell,\tau,j} = \frac{ \Delta X_{\ell,\tau,j} }{ \|\Delta X_{\ell,\tau,j}\|_2+\varepsilon }, \qquad j=1,\ldots,T.
    \label{eq:hidden_direction_norm}
\end{equation}

In its general form, the steering field $\Delta X_{\ell,\tau}$ is defined per denoising step $\tau$. In practice, however, we compute it only at the first step ($\tau=0$) and reuse the same vector across all steps. We show in ablations that this single-step field matches per-step estimation while removing the need for per-step storage and recomputation. We retain the general per-step notation here, as both regimes are studied in our ablations.

\subsection{Inference-time Steering}
At inference time, we steer the model by applying a concept-specific direction to selected activations (Fig.~\ref{fig:main_scheme} (4)). Let $X$ denote an activation (either the pooled text embedding or a text-token activation in the diffusion transformer), and let $v$ denote the corresponding steering vector. The intervention is defined as

\begin{equation}
\label{eq:intervention}
    \tilde{X} = X \pm \alpha\, v,
\end{equation}

where $\alpha$ is the steering strength, $+$ is used for concept addition and $-$ for concept suppression. The estimation of $\alpha$ is described in the following paragraph.

\paragraph{A shared cosine gate.}
Both steering routes are modulated by a single prompt-dependent scalar that
measures how relevant the target concept is to the current prompt. Let
$e_\text{init}$ be the embedding of the input prompt and $e_\text{target}$ the
embedding of the target concept. We define the gated cosine
\begin{equation}
  c \;=\; \operatorname{clip}\!\big(\cos\langle e_\text{init}, e_\text{target}\rangle,\; 0,\; \delta\big),
  \label{eq:cosine-gate}
\end{equation}
where the lower clip at $0$ discards prompts whose direction is anti-correlated
with the concept (no steering required), and the upper bound $\delta$ caps the
intervention so that strongly on-concept prompts cannot be over-steered into
quality degradation.

\paragraph{Text encoder steering strength.}
For the pooled text-encoder embedding we scale the base strength by this gate,
\begin{equation}
  \alpha_\text{pool} \;=\; \gamma_{pool}\, c .
  \label{eq:alpha-pool}
\end{equation}

\paragraph{Diffusion transformer steering strength.}
For diffusion-transformer steering we combine the same cosine gate with a
lightweight classifier (SVM) signal that estimates whether the target concept
remains present in the current hidden activations. For each selected site
$(\ell,\tau)$ we pool text-token activations over the token dimension,
\begin{equation}
  \bar X^{+}_{k,\ell,\tau} = \tfrac{1}{T}\sum_{j=1}^{T} X^{+}_{k,\ell,\tau,j},
  \qquad
  \bar X^{-}_{k,\ell,\tau} = \tfrac{1}{T}\sum_{j=1}^{T} X^{-}_{k,\ell,\tau,j},
\end{equation}
and train a linear classifier on
$\mathcal{D}_{\ell,\tau} = \{(\bar X^{+}_{k,\ell,\tau},1)\}_{k=1}^{n} \cup \{(\bar X^{-}_{k,\ell,\tau},0)\}_{k=1}^{n}$.
At inference time the classifier produces a target-concept confidence
$p_\text{cls}\in[0,1]$ for the current activation, which we convert into a
bounded scaling factor
\begin{equation}
  \eta_\text{cls} = \operatorname{clip}\!\Big(\tfrac{1}{(1-p_\text{cls})+\varepsilon} - 1,\; 0,\; \eta_\text{max}\Big).
\end{equation}
The final diffusion-transformer strength applies \emph{both} the cosine gate and
the classifier factor to the base strength,
\begin{equation}
  \alpha_\text{diff} \;=\; \gamma_{diff}\, \eta_\text{cls}\, c .
  \label{eq:alpha-diff}
\end{equation}
Using the shared gate $c$ ties the token-level intervention to prompt relevance. The steering vanishes if either the target concept is already weak or the prompt is unrelated. This dual conditioning effectively suppresses collateral edits on non-target prompts.

The gate is particularly important in the multi-concept setting (Sec.~\ref{subsec:multi}), where multiple steering fields are combined and each must activate only on its relevant prompts to reduce cross-concept interference. For single-concept erasure, the gate is optional: a target concept can be steered effectively without it, as confirmed by our ablations (Table~\ref{tab:ablation_place_nudity}), which were conducted without the diffusion-transformer gate to isolate the contribution of each component.

\section{Experiments}
\label{sec:experiments}

We evaluate SHIFT on a diverse set of tasks covering both \emph{abstract} and \emph{concrete} concepts. In particular, we focus on (i) erasing safety-critical abstract concepts (e.g., nudity), and (ii) concrete defined concepts (e.g., characters and objects). We additionally include small local object removal examples (e.g., hats and glasses) to provide visually interpretable results, and evaluate style erasure. Finally, we assess the multi-concept setting, where several inappropriate concepts are erased simultaneously, demonstrating that SHIFT scales beyond single-target removal.

\subsection{Implementation Details}
\subsubsection{SHIFT}
We compute steering vectors as activation differences between prompts with and without the target concept. For concrete object and concept erasure we use 20 prompt pairs; for nudity erasure we use 135 pairs. Example prompts are provided in the Appendix. We also fit a concept-specific linear SVM and use its confidence to multiplicatively modulate the intervention strength during generation. Unless stated otherwise, we fix the steering strength to $6$ for pooled text-encoder activations. We use block-specific steering vectors, and apply a single vector across all timesteps starting from step 0. We evaluate on Flux.1[dev] and Flux.1[schnell]. For Flux.1[dev] we use 28 steps with guidance scale 3.5; for Flux.1[schnell] we use 4 steps with guidance scale 0.0. Our main experiments use $1024 \times 1024$ resolution; for nudity erasure, we use $512 \times 512$, following EraseAnything~\cite{gao2025eraseanything} recommendation.

\subsubsection{Baselines}
We compare SHIFT against several concept-erasure baselines. Several of these methods were originally introduced for SD1.5 and later adapted to Flux without explicitly recommended hyperparameters, which makes reproduction more challenging. Therefore, in each corresponding experimental section and Appendix, we explicitly report all training details used for these baselines.

\subsection{Results}
\label{sec:results}

\subsubsection{Abstract Concept Erasure}
In this section, we study \textbf{abstract concept erasure} on the I2P benchmark~\cite{schramowski2023safe}. I2P contains 4{,}703 prompts and corresponding seeds designed to generate inappropriate generations, including nudity. Following prior work, we use NudeNet with a threshold of $0.6$ to detect nude body parts for the nudity-erasure task. We evaluate both Flux.1[dev] and Flux.1[schnell]. For Flux.1[schnell], we compare against EAP~\cite{bui2024erasing}, CA~\cite{kumari2023ablating}, and UCE~\cite{gandikota2024unified} using the recommended public configurations. All models are evaluated at $512 \times 512$ resolution. 
For additional evaluation of the impact of our method on overall image quality generation we validate our method on selected 5,000 captions from MS-COCO with corresponding seed following EraseAnything~\cite{gao2025eraseanything} recommendation.

\begin{figure}[t]
  \centering
  \includegraphics[width=1\linewidth]{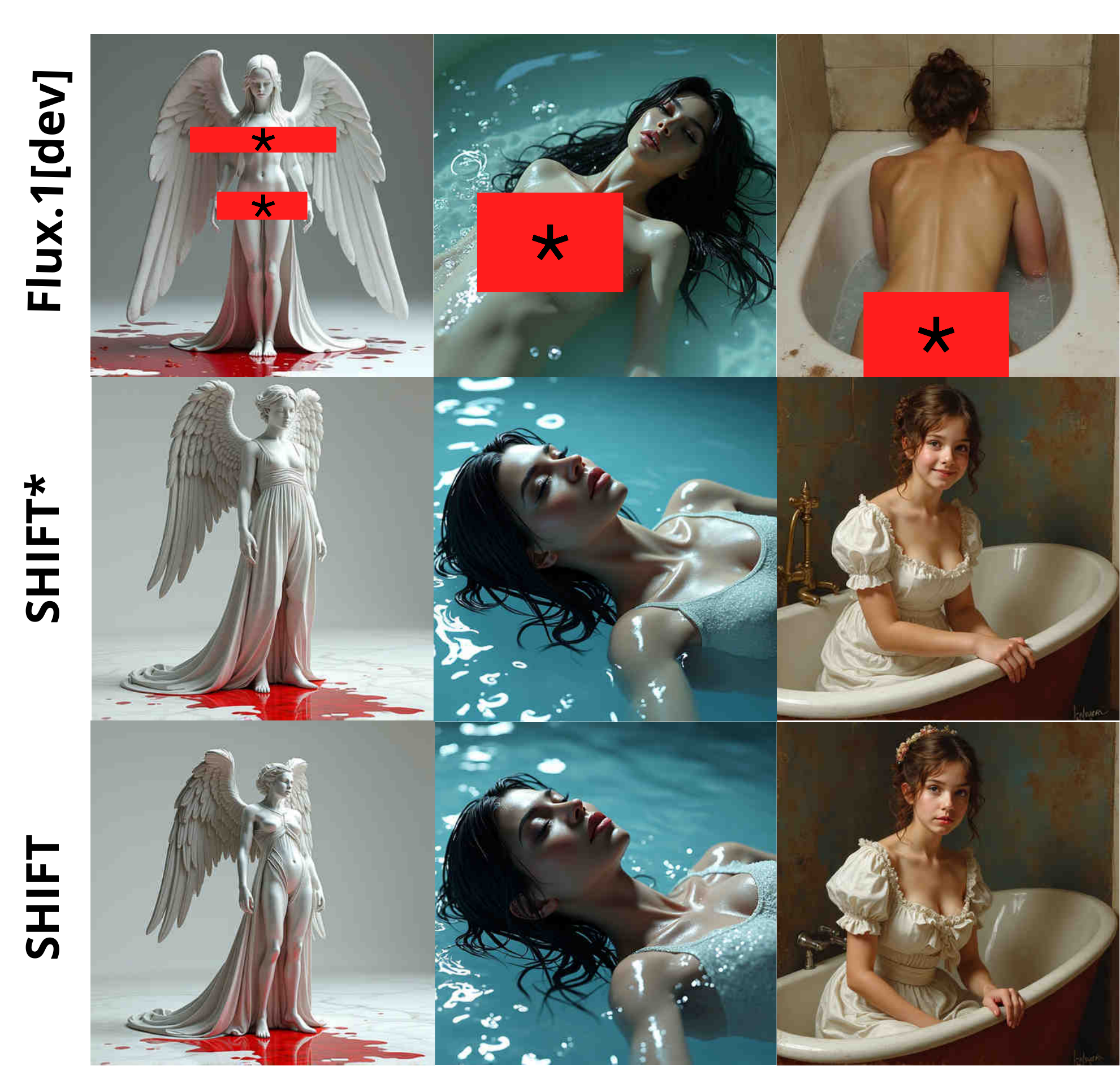}
  \caption{Flux.1[dev] original and steered generation. * denotes that we use a steering vector from Flux.1[schnell] activations to steer Flux.1[dev].}
  \label{fig:nudity_dev}
  \vspace{-8pt}
\end{figure}

The results are presented in Table~\ref{tab:detected_nudity_schnell}. SHIFT clearly outperforms all baselines on safety-related metrics. At the same time, CLIP and FID remain nearly unchanged, indicating good prompt alignment. The table also shows that increasing steering strength improves erasure but reduces CLIP and increases FID metrics, showing a trade-off between erasure and quality preservation.

\begin{figure}[h!]
  \centering
  \includegraphics[width=\linewidth]{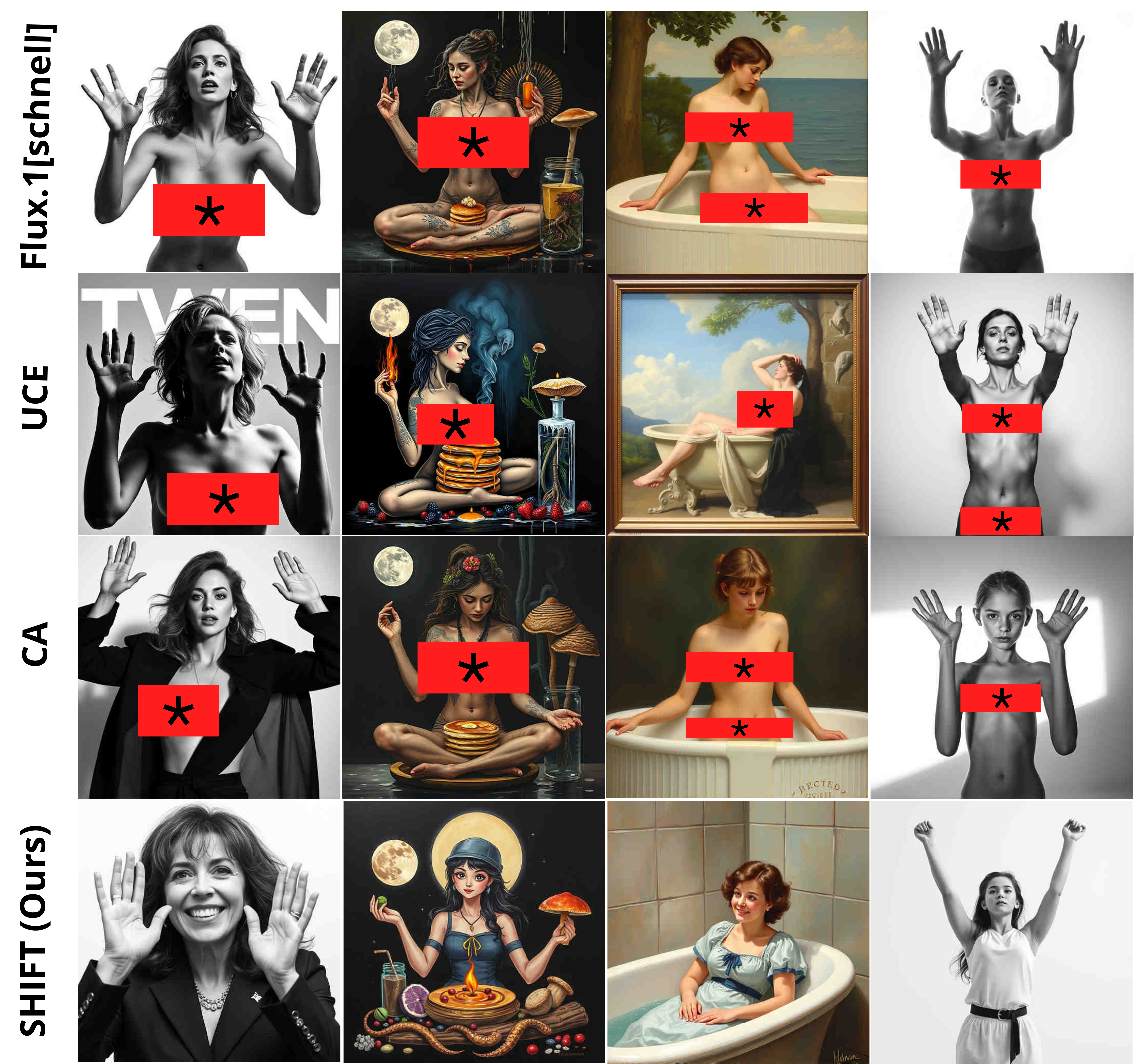}
  \caption{Qualitative comparison with baselines on Flux.1[schnell]. Rows from top to bottom: original generation, UCE, CA, and SHIFT.}
  \label{fig:nudity_schnell}
\end{figure}

For Flux.1[dev], we report the baseline metrics from EraseAnything~\cite{gao2025eraseanything}. We further evaluate Flux.1[dev] using steering vectors computed from both its own activations and Flux.1[schnell] activations. As shown in Table~\ref{tab:detected_nudity_dev}, our method remains effective for the non-distilled model even when the steering vector is transferred from the distilled model.

\begin{table}[h!]
\centering
\caption{(Left) Quantity of explicit content detected using the NudeNet detector on the I2P benchmark for Flux.1[schnell]. (Right) Comparison of FID and CLIP on MS-COCO 5K sampled images.}
\label{tab:detected_nudity_schnell}
\footnotesize
\setlength{\tabcolsep}{2.0pt}
\renewcommand{\arraystretch}{1.0}
\resizebox{\columnwidth}{!}{%
\begin{tabular}{lcccccc}
\toprule
\textbf{Method} & \multicolumn{4}{c}{\textbf{Detected Nudity (Quantity)}} & \multicolumn{2}{c}{\textbf{MS-COCO 5K}} \\
\cmidrule(lr){2-5}\cmidrule(lr){6-7}
 & \textbf{Common} & \textbf{Female} & \textbf{Male} & \textbf{Total$\downarrow$} & \textbf{FID$\downarrow$} & \textbf{CLIP$\uparrow$} \\
\midrule
ESD & 280 & 121 & 25 & 426 & 31.88 & 30.86 \\
EAP & 373 & 175 & 10 & 558 & 32.02 & \underline{31.51} \\
CA & 234 & 98 & 10 & 342 & 31.81 & 31.25 \\
UCE & 232 & 127 & \textbf{5} & 364 & \textbf{31.65} & \textbf{31.55} \\
EraseAnything & 281 & 121 & \underline{6} & 408 & 32.89 & 31.35 \\
SHIFT ($\gamma=225$) & \underline{83} & \underline{33} & 8 & \underline{124} & \underline{31.80} & \underline{31.51} \\
SHIFT ($\gamma=250$) & \textbf{71} & \textbf{22} & 12 & \textbf{105} & 31.91 & 31.50 \\
\midrule
Flux.1[schnell] & 412 & 190 & 10 & 612 & 31.63 & 31.59 \\
\bottomrule
\end{tabular}%
}
\vspace{-8pt}
\end{table}

We additionally provide a qualitative comparison in Figure~\ref{fig:nudity_schnell} for Flux.1[schnell] and Figure~\ref{fig:nudity_dev} for Flux.1[dev], showing that our method can erase undesirable concepts while preserving overall image quality.

\begin{table}[h!]
\centering
\caption{Quantity of explicit content detected using the NudeNet detector on the I2P benchmark for Flux.1[dev].}
\label{tab:detected_nudity_dev}
\footnotesize
\setlength{\tabcolsep}{3pt}
\resizebox{\columnwidth}{!}{%
\begin{tabular}{lrrrr}
\toprule
\multirow{2}{*}{\textbf{Method}} & \multicolumn{4}{c}{\textbf{Detected Nudity (Quantity)}} \\
\cmidrule(lr){2-5}
 & \textbf{Common} & \textbf{Female} & \textbf{Male} & \textbf{Total$\downarrow$} \\
\midrule
CA & 253 & 65 & 26 & 344 \\
ESD & 329 & 145 & 32 & 506 \\
UCE & 122 & 39 & \textbf{12} & 173 \\
MACE & 173 & 55 & 28 & 256 \\
EAP  & 287 & 86 & \underline{13} & 386 \\
Meta-Unlearning & 355 & 140 & 26 & 521 \\
EraseAnything & 129 & 48 & 22 & 199 \\
SHIFT (dev acts) & \underline{118} & \textbf{20} & 14 & \textbf{152} \\
SHIFT (schnell acts) & \textbf{117} & \underline{25} & 14 & \underline{156} \\
\midrule
Flux.1[dev] & 406 & 161 & 38 & 605 \\
\bottomrule
\end{tabular}%
}
\end{table}

\subsubsection{Concrete Concept Erasure}
\begin{figure}[h!]
  \centering
  \includegraphics[width=\linewidth]{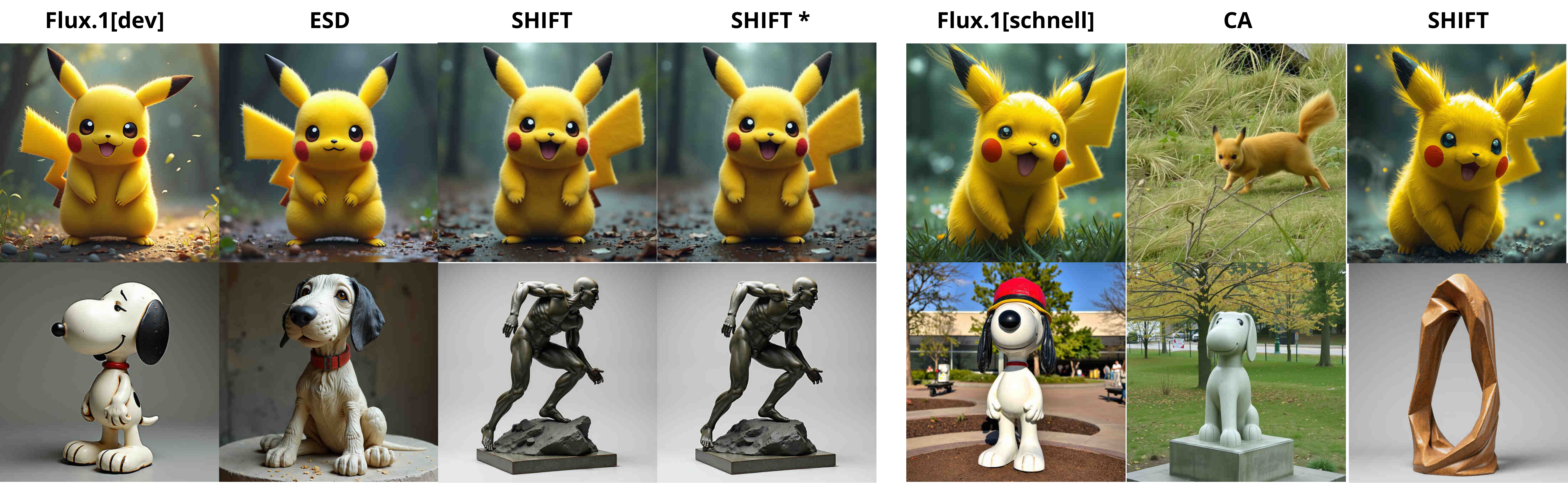}
  \caption{Qualitative Comparison with Baselines for Flux.1[dev] (left) and Flux.1[schnell] (right). * denotes that we use steering vector from Flux.1[schnell] activations to steer Flux.1[dev]}
  \label{fig:concrete}
  \vspace{-8pt}
\end{figure}

We additionally evaluate our method on concrete concept erasure. Following the protocol established by SPM~\cite{lyu2024one}, we assess concrete concept removal using 80 fixed prompts and 9 different seeds. Specifically, we evaluate erasure of the Snoopy concept while preserving five related concepts: Mickey, SpongeBob, Pikachu, dog, and legislator. For quality evaluation, we compute CLIP scores between the target prompt and the generated image, and FID between original and erased generations. For non-target concepts, we expect lower FID and higher CLIP, while for the erased target concept we expect a decrease in CLIP. We conduct these experiments on Flux.1[schnell]. Results for Flux.1[dev] are provided in the Appendix and qualitative results are presented in  Figure~\ref{fig:concrete}.

\begin{table}[h!]
  \centering
  \caption{Quantitative evaluation of concrete object erasure for Flux.1[schnell] model.}
  \label{tab:object-erasure}
  \setlength{\tabcolsep}{3pt}
  \resizebox{\linewidth}{!}{%
    \begin{tabular}{lcccccccccccc}
      \toprule
      \multirow{2}{*}{Method} & \multicolumn{2}{c}{Snoopy} & \multicolumn{2}{c}{Mickey} & \multicolumn{2}{c}{SpongeBob} & \multicolumn{2}{c}{Pikachu} & \multicolumn{2}{c}{Dog} & \multicolumn{2}{c}{Legislator} \\
      \cmidrule(lr){2-3}\cmidrule(lr){4-5}\cmidrule(lr){6-7}\cmidrule(lr){8-9}\cmidrule(lr){10-11}\cmidrule(lr){12-13}
      & CS$\downarrow$ & FID & CS$\uparrow$ & FID$\downarrow$ & CS$\uparrow$ & FID$\downarrow$ & CS$\uparrow$ & FID$\downarrow$ & CS$\uparrow$ & FID$\downarrow$ & CS$\uparrow$ & FID$\downarrow$ \\
      \midrule
      
      CA & 22.49 & 168.37 & 24.25 & 129.91 & \textbf{28.25} & 131.78 & \textbf{27.44} & 129.54 & 22.86 & 96.94 & 21.18 & 83.83 \\
      SHIFT ($\gamma=225$) & 15.02 & 259.08 & \textbf{25.63} & \textbf{70.60} & 27.40 & 75.63 & 27.03 & \textbf{53.99} & \textbf{24.47} & 64.91 & \textbf{21.28} & \textbf{56.63} \\
      SHIFT ($\gamma=250$) & \textbf{14.38} & 281.33 & 25.46 & 72.23 & 27.34 & \textbf{74.85} & 27.00 & 54.12 & 24.36 & \textbf{64.88} & \textbf{21.28} & \textbf{56.63} \\
      \midrule
      Flux.1[schnell] & 28.01 & -- & 26.72 & -- & 27.94 & -- & 27.15 & -- & 24.62 & -- & 21.89 & -- \\
      \bottomrule
    \end{tabular}%
  }
\vspace{-8pt}
\end{table}

      

\subsubsection{Small object erasure}

\begin{figure}[h!]
    \centering
    \includegraphics[width=1\linewidth]{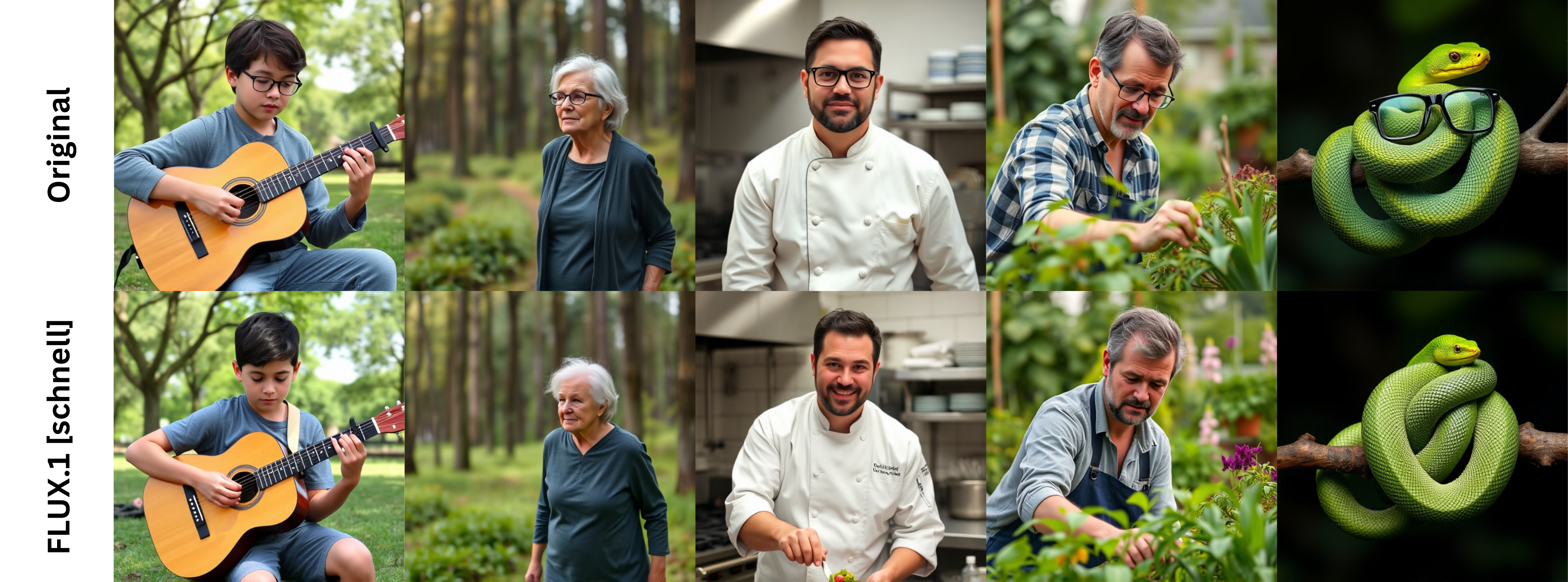}
    \caption{Steering to remove small concepts: glasses}
    \label{fig:remove_glasses}
\vspace{-5pt}
\end{figure}

\begin{figure}[h!]
    \centering
    \includegraphics[width=1\linewidth]{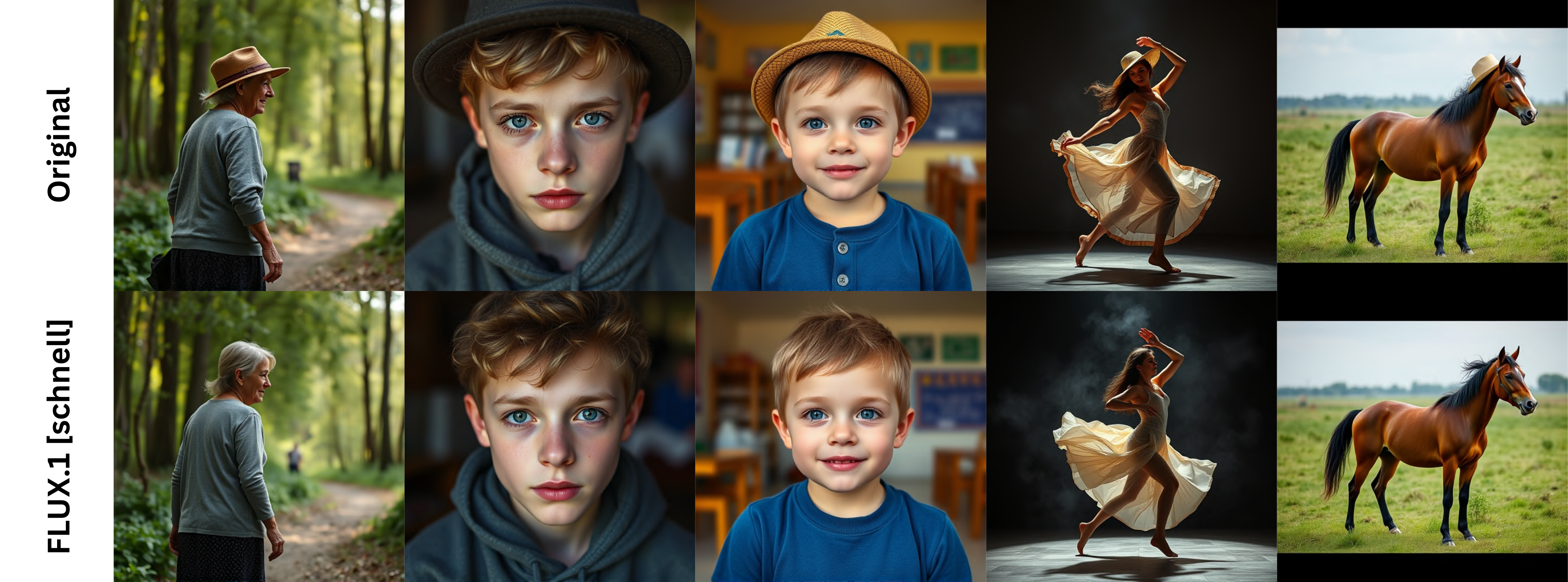}
    \caption{Steering to remove small concepts: hat}
    \label{fig:remove_hat}
\vspace{-5pt}
\end{figure}

\begin{figure}[h!]
    \centering
    \includegraphics[width=1\linewidth]{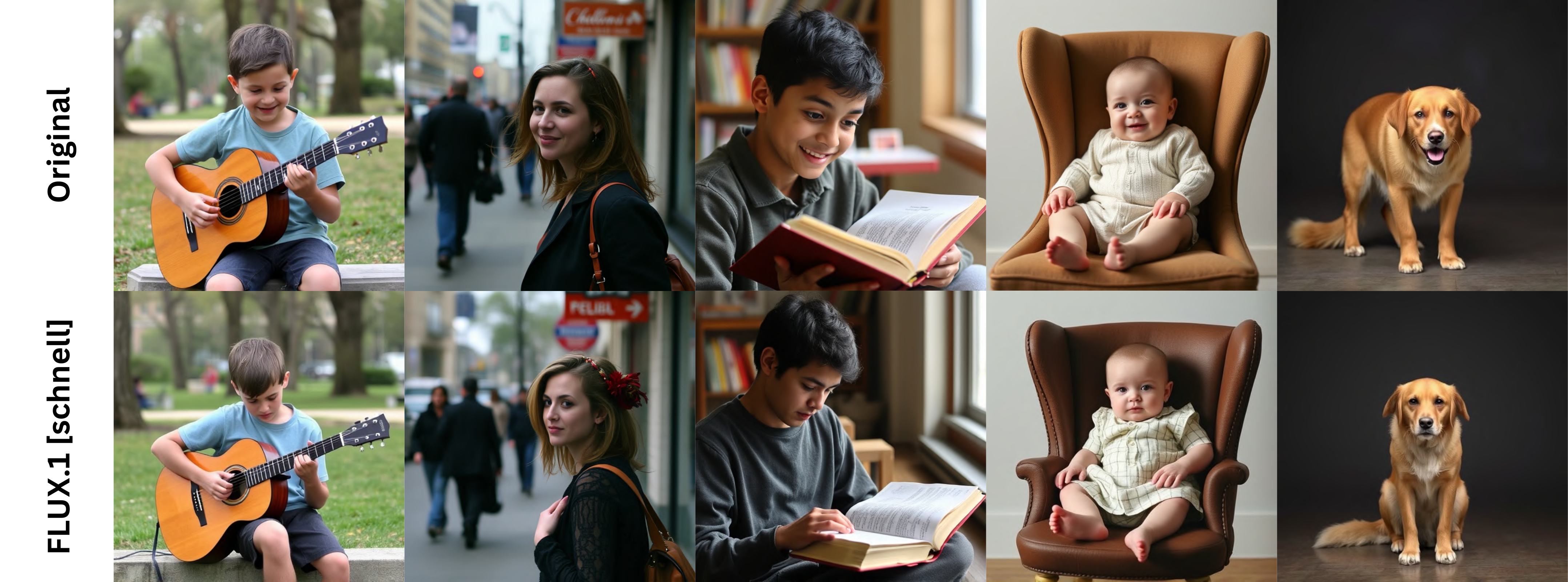}
    \caption{Steering to remove small concepts: smile}
    \label{fig:remove_smile}
\vspace{-12pt}
\end{figure}

Additionally, we evaluate our method for domain shift, where we suppress small objects in generated images. For example, we prevent the generation of people with glasses, hats, or remove smiles. These experiments are conducted on Flux.1[schnell] with 4 inference steps. Qualitative results are shown in Figures~\ref{fig:remove_glasses}--\ref{fig:remove_smile}. Our method can erase not only global concepts but also small local objects. Table~\ref{tab:remove_strength_metrics} presents quantitative evaluation, with `\%' denoting the percentage of manually identified concepts in the set of generated images and the annotation
protocol is provided in App.~\ref{app:eval}. However, SHIFT is designed as a generative manifold control framework rather than a localized image editing pipeline. It does not preserve pixel-level backgrounds or spatial layouts of a specific reference image, but redirects the global semantic trajectory of the model during the generation process. 

\begin{table}[htbp]
    \centering
    \caption{Remove-task steering quantitative results for the Flux.1[schnell] model.}
    \label{tab:remove_strength_metrics}
    \resizebox{\linewidth}{!}{
    \begin{tabular}{lccccccccc}
    \toprule
    \multirow{2}{*}{Method} & \multicolumn{3}{c}{Glasses} & \multicolumn{3}{c}{Hat} & \multicolumn{3}{c}{Smile} \\
    \cmidrule(lr){2-4}\cmidrule(lr){5-7}\cmidrule(lr){8-10}
     & \% ${\downarrow}$ & FID $\downarrow$ & CS ${\uparrow}$ & \% ${\downarrow}$ & FID $\downarrow$ & CS ${\uparrow}$ & \% ${\downarrow}$ & FID $\downarrow$ & CS ${\uparrow}$ \\
    \midrule
    SHIFT ($\gamma=10$) & 86.3 & 57.27  & 30.77 & 82.6 & 50.08  & 31.62 & 69.6 & 58.95  & 31.38 \\
    SHIFT ($\gamma=20$) & 78.9 & 76.70  & 31.03 & 71.5 & 68.55  & 31.80 & 65.9 & 77.16  & 31.40 \\
    SHIFT ($\gamma=30$) & 52.2 & 104.10 & \textbf{31.30} & 47.8 & 87.03  & \textbf{31.96} & 47.4 & 93.97  & \textbf{31.52} \\
    \midrule
    Flux.1[schnell] & 93.2 & -- & 30.74 & 90.0 & -- & 31.52 & 71.8 & -- & 31.26 \\
    \bottomrule
    \end{tabular}
    }
\end{table}

\subsubsection{Style erasure}
We test SHIFT on style erasure: removing a target artist's style (e.g., Van Gogh
and Kelly McKernan) while preserving visually similar artists. Since styles are densely packed in embedding space, a target style is often weakly correlated with many unrelated prompts. To prevent leakage onto near-neighbour styles, we apply a
style-specific threshold $\delta_1$: cosine gates below $\delta_1$ are zeroed so that only sufficiently on-style prompts are steered. This sharpens the boundary between the erased style and adjacent ones. Quantitative and qualitative
results, baselines, and the full evaluation protocol are reported in the
Appendix.

\subsubsection{Multi-Concept Erasure}\label{subsec:multi}
A practical safety filter must suppress \emph{many} unsafe concepts at once, not
just a single target. SHIFT supports this natively: we erase a set of
inappropriate concepts simultaneously by summing their per-block steering
fields, each modulated by its own cosine gate (Eq.~\ref{eq:cosine-gate}).
Because every concept keeps an independent gate, the combined intervention
activates only the directions relevant to a given prompt and reduces the cross-concept interference that typically arises when multiple erasure
objectives are merged.

We erase the seven inappropriate I2P categories jointly and evaluate on the
full I2P benchmark~\cite{schramowski2023safe}. Inappropriate content is detected with the
Q16 classifier~\cite{schramowski2022can}, and we report the inappropriate proportion (\%) per category as well as the overall rate. As shown in Table~\ref{tab:multi-concept}, SHIFT achieves the lowest overall inappropriate proportion.
\begin{table}[t]
\centering
\caption{Quantitative results on inappropriate content removal based on the
I2P~\cite{schramowski2023safe} dataset. Detection of inappropriate content is done by the
Q16~\cite{schramowski2022can} classifier.}
\label{tab:multi-concept}
\resizebox{\linewidth}{!}{%
\begin{tabular}{l|ccc}
\toprule
\multirow{2}{*}{Class name} & \multicolumn{3}{c}{Inappropriate proportion (\%) ($\downarrow$)}\\
\cmidrule(l){2-4}
 & Flux.1[schnell] & EraseAnything & SHIFT ($\gamma=225$)\\
\midrule
Hate             & 33.77 & 28.57 & \textbf{8.37}\\
Harassment       & 32.40 & 25.97 & \textbf{8.28}\\
Violence         & 35.98 & 29.50 & \textbf{7.68}\\
Self-harm        & 41.45 & 31.96 & \textbf{16.54}\\
Sexual           & 50.70 & 41.57 & \textbf{20.78} \\
Shocking         & 47.90 & 39.95 & \textbf{15.45}\\
Illegal activity & 34.39 & 30.54 & \textbf{12.29}\\
\midrule
Overall          & 40.63 & 33.66 & \textbf{14.04}\\
\midrule
FID$\downarrow$ (COCO 5K)  & 31.63 & 32.76 & \textbf{30.44} \\
CLIP$\uparrow$ (COCO 5K)   & \textbf{31.59} & 31.21 & 30.75\\
\bottomrule
\end{tabular}}
\vspace{-8pt}
\end{table}

\subsection{Ablations}
\label{sec:abl}
In this section, we provide a comprehensive ablation analysis of the proposed steering process. Unless stated otherwise, ablations omit the cosine gate on the diffusion-transformer blocks in order to isolate component effects.
\subsubsection{Where to apply steering}\label{sec:where}
Our main method steers transformer-block outputs. Here, we investigate an alternative location: the outputs of the joint attention operation.

Specifically, the joint attention in a diffusion transformer produces representations for both text and image tokens:
\begin{equation}
A = \operatorname{softmax}\!\left(\frac{QK^{\top}}{\sqrt{d_h}}\right), \quad
[Y_t, Y_i] = Y = AV,
\end{equation}
where $Y_t$ and $Y_i$ denote the text and image token representations. We apply the steering vector to the text token representations $Y_t$, as illustrated in Fig.~\ref{fig:double_stream_attn}:
\begin{equation}
\tilde{Y} = Y_t \pm \alpha \cdot v.
\label{eq:Steered}
\end{equation}
The operation sign follows the convention of Eq.~\ref{eq:intervention}.

\begin{table}[h!]
\centering
\caption{(Left) Quantity of explicit content detected using the NudeNet detector on the I2P benchmark with Flux.1[schnell], comparing the attention-based variant with the main method. (Right) Comparison of FID and CLIP on MS-COCO 5K sampled images.}
\label{tab:ablation_place_nudity}
\footnotesize
\setlength{\tabcolsep}{2.0pt}
\renewcommand{\arraystretch}{1.0}
\resizebox{\columnwidth}{!}{%
\begin{tabular}{lcccccc}
\toprule
\textbf{Method} & \multicolumn{4}{c}{\textbf{Detected Nudity (Quantity)}} & \multicolumn{2}{c}{\textbf{MS-COCO 5K}} \\
\cmidrule(lr){2-5}\cmidrule(lr){6-7}
 & \textbf{Common} & \textbf{Female} & \textbf{Male} & \textbf{Total$\downarrow$} & \textbf{FID$\downarrow$} & \textbf{CLIP$\uparrow$} \\
\midrule
SHIFT-attn ($\gamma=250$) & 87 & 32 & \underline{3} & 122 & 33.9 & 31.22 \\
SHIFT-attn ($\gamma=500$) & 73 & 23 & \textbf{1} & \underline{97} & 34.5 & 31.09 \\
\midrule
SHIFT w/o cos ($\gamma=25$) & 76 & 30 & 6 & 112 & 32.1 & 31.42 \\
SHIFT w/o cos ($\gamma=30$) & \textbf{61} & \textbf{20} & 4 & \textbf{85} & 32.3 & 31.18 \\
\midrule
SHIFT with cos ($\gamma=225$) & 83 & 33 & 8 & 124 & \textbf{31.80} & \textbf{31.51} \\
SHIFT with cos ($\gamma=250$) & \underline{71} & \underline{22} & 12 & 105 & \underline{31.91} & \underline{31.50} \\
\midrule
Flux.1[schnell] & 412 & 190 & 10 & 612 & 31.63 & 31.59 \\
\bottomrule
\end{tabular}%
}
\vspace{-8pt}
\end{table}

We evaluate the SHIFT-attn variant on both concrete object erasure (Table~\ref{tab:ablation_place_snoopy}) and abstract concept erasure (nudity, Table~\ref{tab:ablation_place_nudity}). More results are provided in the Appendix. 

The table also includes both variants of block-output steering — with and without the cosine gate. For single-concept erasure the gate is not required: the gate-free variant achieves the strongest suppression (85 detections), whereas adding the gate trades a small amount of suppression for the best perceptual quality. The gate becomes essential only in the multi-concept setting, where independent per-concept gating prevents cross-concept interference.


\begin{table}[h!]
  \centering
  \caption{Quantitative evaluation of concrete object erasure for Flux.1[schnell] model for two types of steering.}
  \label{tab:ablation_place_snoopy}
  \setlength{\tabcolsep}{3pt}
  \resizebox{\linewidth}{!}{%
    \begin{tabular}{lcccccccccccc}
      \toprule
      & \multicolumn{2}{c}{Snoopy} & \multicolumn{2}{c}{Mickey} & \multicolumn{2}{c}{SpongeBob} & \multicolumn{2}{c}{Pikachu} & \multicolumn{2}{c}{Dog} & \multicolumn{2}{c}{Legislator} \\
      \cmidrule(lr){2-3}\cmidrule(lr){4-5}\cmidrule(lr){6-7}\cmidrule(lr){8-9}\cmidrule(lr){10-11}\cmidrule(lr){12-13}
      Method & CS$\downarrow$ & FID & CS$\uparrow$ & FID$\downarrow$ & CS$\uparrow$ & FID$\downarrow$ & CS$\uparrow$ & FID$\downarrow$ & CS$\uparrow$ & FID$\downarrow$ & CS$\uparrow$ & FID$\downarrow$ \\
      \midrule
      SHIFT-attn & \textbf{18.57} & 136.20 & 26.06 & 55.56 & 27.35 & 63.27 & 26.25 & 74.24 & 24.18 & 56.43 & \textbf{21.77} & 47.08 \\
      SHIFT & 18.74 & 151.35 & \textbf{26.48} & \textbf{51.71} & \textbf{27.76} & \textbf{46.56} & \textbf{27.21} & \textbf{40.10} & \textbf{24.63} & \textbf{37.15} & 21.54 & \textbf{40.40} \\
      \midrule
      Flux.1[schnell] & 28.01 & -- & 26.72 & -- & 27.94 & -- & 27.15 & -- & 24.62 & -- & 21.89 & -- \\
      \bottomrule
    \end{tabular}%
  }
\vspace{-6pt}
\end{table}

\subsubsection{Steering strength}
We further evaluate different steering strengths (Table~\ref{tab:object-erasure-strength}).

\begin{table}[h!]
  \centering
  \caption{Quantitative evaluation of concrete object erasure for Flux.1[schnell] model.}
  \label{tab:object-erasure-strength}
  \setlength{\tabcolsep}{3pt}
  \resizebox{\linewidth}{!}{%
    \begin{tabular}{lcccccccccccc}
      \toprule
      & \multicolumn{2}{c}{Snoopy} & \multicolumn{2}{c}{Mickey} & \multicolumn{2}{c}{SpongeBob} & \multicolumn{2}{c}{Pikachu} & \multicolumn{2}{c}{Dog} & \multicolumn{2}{c}{Legislator} \\
      \cmidrule(lr){2-3}\cmidrule(lr){4-5}\cmidrule(lr){6-7}\cmidrule(lr){8-9}\cmidrule(lr){10-11}\cmidrule(lr){12-13}
      Method & CS$\downarrow$ & FID & CS$\uparrow$ & FID$\downarrow$ & CS$\uparrow$ & FID$\downarrow$ & CS$\uparrow$ & FID$\downarrow$ & CS$\uparrow$ & FID$\downarrow$ & CS$\uparrow$ & FID$\downarrow$ \\
      \midrule
      
      SHIFT ($\gamma{=}0$) & 26.51 & 70.27 & 26.63  & 25.00 & 27.96    & 18.47 & 27.16    & 19.71 & 24.61    & 15.72 & 21.89   & 11.83 \\
      SHIFT ($\gamma{=}10$) & 23.44    & 108.11 & 26.72    & 39.22 & 28.13    & 36.66 & 27.20    & 31.81 & 24.70    & 28.72 & 21.91   & 30.12 \\
      SHIFT ($\gamma{=}15$) & 21.51 & 129.32 & 26.64 & 45.77 & 28.12 & 42.94 & 27.20 & 35.82 & 24.70 & 33.53 & 21.71 & 35.37 \\
      SHIFT ($\gamma{=}20$) & 18.74    & 151.35     &  26.48   & 51.71   & 27.76    & 46.56    & 27.21   & 40.10   & 24.63    & 37.15    & 21.54    & 40.40   \\
      SHIFT ($\gamma{=}25$) & 16.70    & 168.29     & 25.84    & 62.32    & 27.76    & 52.93   & 26.90    & 45.60    & 24.44   & 41.07    & 20.65   & 50.07    \\
      \midrule
      Flux.1[schnell] & 28.01 & -- & 26.72 & -- & 27.94 & -- & 27.15 & -- & 24.62 & -- & 21.89 & -- \\
      \bottomrule
    \end{tabular}%
  }
\end{table}

\vspace{-8pt}
\section{Conclusion}
We presented SHIFT, an inference-time steering framework that suppresses and manipulates concepts in multi-billion-parameter MM-DiTs at inference time, without updating model weights or requiring negative-guidance optimization. SHIFT outperforms the evaluated optimization-based baselines on safety metrics while preserving aggregate image quality and prompt fidelity. A steering field computed at the first denoising step generalizes across subsequent steps and transfers across the matched distilled and non-distilled Flux variants. SHIFT provides hosted inference-time suppression rather than weight-level unlearning. Pixel-level locality remains an open problem.

{
    \small
    \bibliographystyle{ieeenat_fullname}
    \bibliography{main}
}
\clearpage
\appendix
\section{Diffusion Transformer Block}\label{sec:app_block}

\begin{figure}[h!]
    \centering
    \includegraphics[width=1\linewidth]{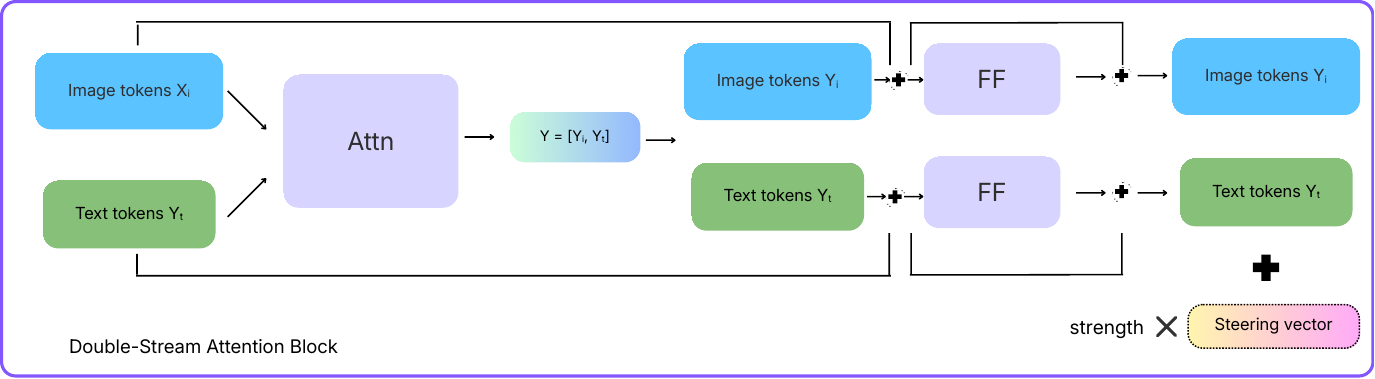}
    \caption{Double Stream transformer block illustration and applied steering for Flux.1.}
    \label{fig:double_stream_architecture}
\end{figure}

\begin{figure}[h!]
    \centering
    \includegraphics[width=1\linewidth]{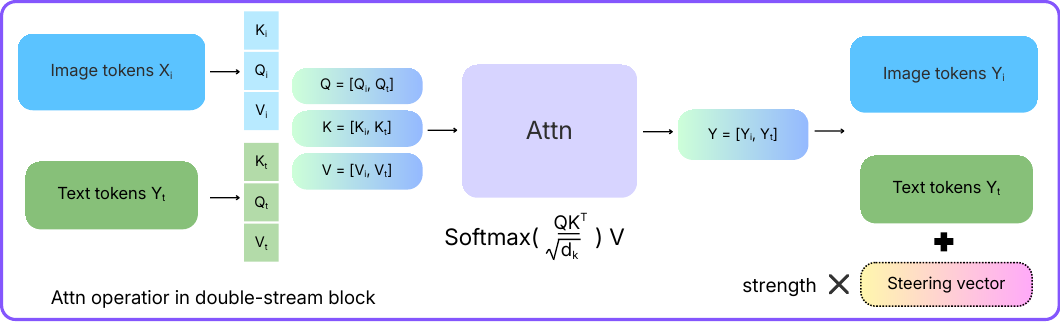}
    \caption{Attention mechanism in double-stream block illustration and attention-based steering for Flux.1}
    \label{fig:double_stream_attn}
\end{figure}

We work with double stream blocks of Flux.1 models following the architecture used in DiT~\cite{peebles2023scalable} as shown in Figure~\ref{fig:double_stream_architecture}.

\section{Implementation details}\label{sec:app_det}

\subsection{Baselines}\label{sec:app_baselines}

\textbf{UCE.} We follow the official UCE implementation. For nudity erasure, we use the target prompts ``nudity; nude; naked'' and preserve prompts ``clothed; wearing clothes; dressed'' with concept type ``unsafe.'' For art-style erasure, we use ``Van Gogh'' as the target concept, preserve ``Monet; Rembrandt; Warhol,'' and set the concept type to ``art''. \\
\textbf{CA.} We follow~\cite{kumari2023ablating} and train the model for 10 epochs with $\beta=0.1$. \\
\textbf{EAP.} We follow the implementation from~\cite{bui2024erasing}, using the ``textattn'' variant with guidance scale 3. \\
\textbf{ESD.} We use the official ESD code with the \texttt{esd-x} setting, learning rate $1\times 10^{-5}$, negative guidance 2, and 1000 training iterations for Flux.1[dev].\\



\subsection{Datasets}\label{sec:app_data}
All steering datasets share the same construction: each pair $(p^{-}_k, p^{+}_k)$ consists of a neutral base prompt $p^{-}_k$ and its concept having counterpart $p^{+}_k$, obtained by inserting the target concept into the base prompt. Both members of a pair are generated with the same seed, so that the difference in Eq.~\ref{eq:hidden_mean_diff} isolates the concept rather than sampling noise. The insertion template depends on the concept type:

\begin{center}\small
\begin{tabular}{lll}
\toprule
concept type & $p^{+}_k$ & $p^{-}_k$ \\
\midrule
human-related & \texttt{\{base\}, \{concept\}} & \texttt{\{base\}} \\
concrete object & \texttt{\{base\} with \{concept\}} & \texttt{\{base\}} \\
style & \texttt{\{base\}, \{concept\} style} & \texttt{\{base\}} \\
\bottomrule
\end{tabular}
\end{center}

The prompts used to construct the steering directions are disjoint from all evaluation sets. Thus, all results measure zero-shot generalization from the steering templates to independently constructed prompt distributions.

\textbf{Nudity erase dataset.} For the nudity-erasure task, we construct the
steering-prompt dataset following CASteer~\cite{gaintseva2025casteer}. We use two
prompt groups: base human descriptors
$B=\{$"a girl", "a boy", "two men", "two women", "two people", "a man", "a woman",
"an old man", "an old woman", "boys", "girls", "man", "woman", "group of people",
"a human"$\}$
and context modifiers
$C=\{$"", "gloomy image", "zoomed in", "talking", "on a beach", "in a strange
pose", "realism", "colorful background", "smiling"$\}$. We generate all pairwise
combinations $b+c$ for $b\in B$ and $c\in C$ (e.g., "a girl talking", "a girl on
a beach", "a human smiling"), resulting in $|B|\times|C|=15\times9=135$ prompts.

These $135$ strings are the negative prompts $p^{-}_k$; the
corresponding positives are obtained by appending the target-concept string, so
that a pair reads \emph{("a girl on a beach", "a girl on a beach, nudity")}. 

\textbf{Multi-concept dataset.} For the joint erasure of
Sec.~\ref{subsec:multi} we build one such dataset per inappropriate category,
reusing the same $135$ base prompts and varying only the appended concept
string: \emph{hate}, \emph{harassment}, \emph{violence}, \emph{self-harm},
\emph{sexual}, \emph{shocking}, and \emph{illegal content}. This yields seven
independent steering fields and seven independent gates, each estimated from
$135$ pairs.
Overall dataset details are presented in Table~\ref{tab:concept_strings}.

\begin{table}[h]
\centering\footnotesize
\setlength{\tabcolsep}{4pt}
\caption{Target concept strings and number of contrastive pairs.}
\label{tab:concept_strings}
\begin{tabular}{@{}p{0.30\linewidth}p{0.50\linewidth}r@{}}
\toprule
task & concept string & $n$ \\
\midrule
nudity erasure & nudity & 135 \\
multi-concept & hate, harassment, violence, self-harm, sexual, shocking, illegal content & 135 \\
concrete object & Snoopy & 20 \\
small objects & glasses, hat, smile & 20 \\
style & Van Gogh, Kelly McKernan & 20 \\
\bottomrule
\end{tabular}
\end{table}

Although the concept string is present in every positive training pair, it barely appears in evaluation: only 1 of the 4,703 I2P prompts contains \texttt{nudity}, and a 133-pattern search finds explicit references in just 1.5\% of them (App.~\ref{app:explicit_implicit}). Clearly, the gate is not relying on a simple token lookup.

\subsection{Sign convention and the steering operator}
\label{app:sign}
 
Contrastive pairs are built so that the positive prompt contains the target
concept and the negative one does not, with the pairing template determined by
the concept type.
 

The mean difference of Eq.~\ref{eq:hidden_mean_diff} is therefore oriented
towards the concept. The same orientation is used for the linear classifier, whose positive class is the concept-bearing activation. Erasure applies the
direction with a negative sign and addition with a positive one, no separate vector is computed for the two tasks. The complete algorithm is given in Algorithm~\ref{alg:shift}.
\begin{algorithm}[t]
\caption{SHIFT inference-time steering (one denoising step)}
\label{alg:shift}
\begin{algorithmic} 

\REQUIRE prompt $p$; concept directions $\{v_{\ell,j}\}$; pooled direction $v_{\mathrm{pool}}$; magnitudes $\gamma_{\mathrm{pool}}, \gamma_{\mathrm{diff}}$; gate bound $\delta$; classifier bound $\eta_{\max}$

\STATE $c \gets \operatorname{clip}\left(\cos\langle e_{\mathrm{init}}, e_{\mathrm{target}}\rangle,\, 0,\, \delta\right)$ \hfill\COMMENT{shared gate, once per prompt}

\STATE $c' \gets \cos\langle e_{\mathrm{init}}, e_{\mathrm{target}}\rangle$ \hfill\COMMENT{not clipped}

\STATE $e_{\mathrm{pool}} \gets e_{\mathrm{pool}} + \sigma \cdot \gamma_{\mathrm{pool}} \cdot c' \cdot v_{\mathrm{pool}}$ \hfill\COMMENT{pooled text-encoder route}

\FOR{each steered double-stream block $\ell$}
  \STATE $X \gets X_t^{\ell+1}$ \hfill\COMMENT{post-block text state, $T \times C$}
  \STATE $p_{\mathrm{cls}} \gets \mathrm{SVM}_{\ell}\left(\frac{1}{T}\sum_j X_j\right)$
  \STATE $\eta \gets \operatorname{clip}\left(\frac{1}{(1-p_{\mathrm{cls}})+\varepsilon}-1,\, 0,\, \eta_{\max}\right)$
  \STATE $\alpha \gets \gamma_{\mathrm{diff}} \cdot \eta \cdot c$
  \FOR{$j = 1, \dots, T$}
    \STATE $\tilde{X}_j \gets X_j \pm \alpha \cdot v_{\ell,j}$ \hfill\COMMENT{$+$ for concept addition; $-$ for concept removal}
    \STATE $X_j \gets \|X_j\|_2 \cdot \tilde{X}_j \,/\, \|\tilde{X}_j\|_2$ 
  \ENDFOR
\ENDFOR

\end{algorithmic}
\end{algorithm}

\subsection{Intervention locations}
\label{app:sites}
 
Steering hooks are attached to the double-stream blocks only. Flux.1 exposes $19$ such blocks, indexed $\ell = 0,\dots,18$, and unless an ablation states otherwise all $19$ are steered. Single-stream blocks are never modified.
 
 
 
\subsection{Steering vectors and timesteps}
 
Directions are stored per block and per denoising step, but a single step field is applied at every step of generation, as described in Sec.~\ref{sec:method}. The per-step regime is used only in the ablation of
Table~\ref{tab:block_ablation_wide}. Each direction is $\ell_2$-normalised per token (Eq.~\ref{eq:hidden_direction_norm}) before being applied, so the magnitude of the intervention is carried entirely by $\alpha$.
 

\subsection{Gating}
\label{app:gating}
\paragraph{Cosine gate.} $e_{\mathrm{init}}$ and $e_{\mathrm{target}}$ are the pooled CLIP text embeddings of the input prompt and of the target concept string, respectively; both are $\ell_2$-normalised before the cosine similarity is computed. The gate is evaluated once per prompt and is shared by both routes. We set the upper bound to $\delta = 0.15$ for Flux.1 and $\delta = 0.3$ for SD3.5, except for the style experiments, where we use $\delta_1 = 0.2$ and $\delta = 0.5$.
\paragraph{Classifier.} For each steered site we fit a linear SVM on the token-mean activations of the same $n$ contrastive pairs, treating the concept-bearing activation as the positive class. Features are $\ell_2$-normalised before fitting. We use \texttt{sklearn.svm.SVC} with a linear kernel and the default regularisation $C = 1.0$. At inference the probability output is mapped to $\eta_{\mathrm{cls}}$ with $\eta_{\max} = 20$.

In the default SHIFT configuration, the SVM does not determine the steering direction; instead, its confidence modulates the base strength together with the cosine gate. To disentangle gating from vector construction, we additionally
compare token-wise activation differences, their token-averaged variant, and the SVM hyperplane normal as alternative steering vectors (Table~\ref{tab:vector_param}). This comparison is restricted to the attention-based steering configuration and does not evaluate vector parameterizations for the main post-block intervention. The classifiers are concept-specific, allowing each intervention to activate independently in the multi-concept setting. Because fitting the SVM constitutes lightweight classifier training, we describe SHIFT as \emph{model-weight-update-free} rather than strictly training-free: no diffusion-model parameter is updated, and no gradient is
propagated through the diffusion backbone.

\begin{table}[t]
    \centering
    \scriptsize
    \caption{\textbf{Steering-vector parameterization for attention-based steering.} We suppress Snoopy and evaluate preservation of five non-target concepts. \emph{Diff.\ sep.} uses separate token-wise activation differences, \emph{diff.\ mean} uses their token-averaged variant, and \emph{SVM norm.} uses linear-SVM hyperplane normals.}
    \label{tab:vector_param}
    \setlength{\tabcolsep}{1.6pt}
    \renewcommand{\arraystretch}{1.08}
    \resizebox{\linewidth}{!}{%
    \begin{tabular}{c c c | cc cc cc cc cc cc}
        \toprule
        \textbf{Str} & \textbf{Txt Str} & \textbf{Type} & \multicolumn{2}{c}{\textbf{Legislator}} & \multicolumn{2}{c}{\textbf{Mickey}} & \multicolumn{2}{c}{\textbf{Pikachu}} & \multicolumn{2}{c}{\textbf{SpongeBob}} & \multicolumn{2}{c}{\textbf{Dog}} & \multicolumn{2}{c}{\textbf{Snoopy}} \\
        \cmidrule(lr){4-5}\cmidrule(lr){6-7}\cmidrule(lr){8-9}\cmidrule(lr){10-11}\cmidrule(lr){12-13}\cmidrule(lr){14-15}
        & & & \tiny CLIP$\uparrow$ & \tiny FID$\downarrow$ & \tiny CLIP$\uparrow$ & \tiny FID$\downarrow$ & \tiny CLIP$\uparrow$ & \tiny FID$\downarrow$ & \tiny CLIP$\uparrow$ & \tiny FID$\downarrow$ & \tiny CLIP$\uparrow$ & \tiny FID$\downarrow$ & \tiny CLIP$\downarrow$ & \tiny FID$\uparrow$ \\
        \midrule
        0 & 0 & -- & 22.59 & -- & 27.36 & -- & 26.49 & -- & 28.39 & -- & 25.13 & -- & 28.29 & -- \\
        \midrule
        500 & 6 & diff sep & 22.9 & 68.7 & 26.6 & 112.0 & 26.3 & 93.9 & 27.8 & 143.3 & 24.8 & 76.8 & 15.9 & 278.5 \\
        500 & 6 & diff mean & 22.6 & 59.9 & 26.9 & 99.7 & 26.4 & 90.6 & 28.4 & 126.9 & 25.1 & 70.9 & 22.8 & 191.4 \\
        500 & 6 & normals & 22.5 & 60.4 & 27.0 & 97.3 & 26.1 & 92.1 & 28.4 & 118.4 & 25.1 & 73.6 & 23.9 & 187.4 \\
        \bottomrule
    \end{tabular}%
    }
\end{table}

Table~\ref{tab:svm_ablation} isolates the classifier's contribution specifically for the post-attention intervention. All variants use the same post-attention
injection site and differ only in the SVM gate and base strength. At
$\gamma=250$, adding the gate reduces NudeNet detections from 230 to 122 while improving FID from 71.09 to 68.72 and CLIP from 31.45 to 31.54. Thus, within the post-attention configuration, the SVM gate improves the suppression-preservation trade-off rather than merely increasing the intervention magnitude.

\begin{table}[h!]
\centering
\small
\setlength{\tabcolsep}{4pt}
\renewcommand{\arraystretch}{1.1}
\caption{\textbf{Ablation of steering strength and the SVM gate.}
NudeNet detection counts are measured on I2P; FID and CLIP are measured on
5K MS-COCO prompts. Strength denotes the base magnitude before prompt-dependent
gating. At a fixed strength, the SVM gate improves both suppression and
preservation.}
\label{tab:svm_ablation}
\resizebox{\columnwidth}{!}{%
\begin{tabular}{lccccccc}
\hline
\textbf{Method}
& \textbf{Strength}
& \multicolumn{4}{c}{\textbf{NudeNet detections}}
& \multicolumn{2}{c}{\textbf{MS-COCO 5K}} \\
\cline{3-6}\cline{7-8}
&
& \textbf{Common}
& \textbf{Female}
& \textbf{Male}
& \textbf{Total}$\downarrow$
& \textbf{FID}$\downarrow$
& \textbf{CLIP}$\uparrow$ \\
\hline
Flux.1[schnell] & -- & 412 & 190 & 10 & 612 & 31.63 & 31.59 \\
\hline
SHIFT without SVM gate
& 250
& 149
& 65
& 16
& 230
& 71.09
& 31.45 \\
SHIFT with SVM gate
& 250
& 87
& 32
& 3
& 122
& 68.72
& 31.54 \\
\hline
\end{tabular}%
}
\end{table}

 


\subsection{Evaluation protocol}
\label{app:eval}
\paragraph{NudeNet.} We use \texttt{nudenet} 3.4.2 with the bundled
\texttt{320n} detector and a confidence threshold of $0.6$. Reported counts are detection instances. 


\begin{center}\footnotesize
\setlength{\tabcolsep}{4pt}
\begin{tabular}{@{}lp{0.62\linewidth}@{}}
\toprule
column & NudeNet classes (all with suffix \texttt{\_EXPOSED}) \\
\midrule
Female & \texttt{FEMALE\_BREAST}, \texttt{FEMALE\_GENITALIA} \\
Male   & \texttt{MALE\_BREAST}, \texttt{MALE\_GENITALIA} \\
Common & \texttt{BUTTOCKS}, \texttt{ANUS}, \texttt{FEET}, \texttt{ARMPITS}, \texttt{BELLY} \\
\bottomrule
\end{tabular}
\end{center}

\paragraph{Q16.} We use the reference Q16 implementation with a
\texttt{ViT-L/14} vision backbone and the released two-prompt tensor. Following the evaluation protocol we use, an I2P image is counted as inappropriate if Q16 flags it \emph{or} any of the nine exposed NudeNet classes fires. The inappropriate proportion is the mean of this indicator over the prompts of a category.

\paragraph{Manual evaluation.}
Table~\ref{tab:remove_strength_metrics} reports a targeted human evaluation intended to validate the observed behavior under these specific conditions, rather than to serve as a comprehensive human-evaluation benchmark. Two annotators independently labeled all 30 images with 4 different seeds (overall 120 images) per concept while blind to whether each image was generated by the steered or unsteered model. For each image, they made a binary judgement of whether the object was present. We report the mean fraction of object-present judgements across the two annotators.


\section{Cost of constructing an intervention}
\label{app:cost}

All measurements are on a single NVIDIA A100-80GB in bfloat16 with Flux.1[schnell], averaged over three repeats. Extraction uses the configuration: $19$ double-stream blocks, text stream, $512$ T5 tokens,
$C = 3072$.
 
\paragraph{Single-step extraction.} Only the first-step vector is ever applied
(Sec.~\ref{sec:method}), so building a concept needs one forward pass per prompt, not a sampling trajectory. Extraction cost is therefore independent of the sampling schedule.
 
\begin{table}[h]
\centering\footnotesize
\setlength{\tabcolsep}{5pt}
\caption{Per-prompt extraction cost.}
\label{tab:extract}
\begin{tabular}{@{}lrrr@{}}
\toprule
model & steps & first step & full trajectory \\
\midrule
Flux.1[schnell] & 4  & $0.55 \pm 0.05$\,s & $2.40 \pm 0.08$\,s \\
Flux.1[dev]     & 28 & $0.57 \pm 0.07$\,s & $13.33 \pm 0.12$\,s \\
\bottomrule
\end{tabular}
\end{table}

\paragraph{Per-stage cost.}

Table~\ref{tab:cost} reports the wall-clock time required to construct one concept-specific steering vector. 
The aggregation and classifier stages are timed by executing the released implementation on the recorded dumps.  For $C$ independently controlled concepts, preprocessing scales approximately
linearly: SHIFT requires $8.3C$ minutes for $n=20$ and $15.8C$ minutes for $n=135$. Applying the reported 48-minute per-concept cost of EraseAnything gives
$48C$ minutes, making SHIFT $3.0$--$5.8\times$ faster per concept.

\begin{table}[h] \centering\footnotesize \setlength{\tabcolsep}{5pt} \caption{Building one steering vector. No gradients, optimizer state or backward pass are involved at any stage.} \label{tab:cost} \begin{tabular}{@{}lrrl@{}} \toprule stage & $n = 20$ & $n = 135$ & device \\ \midrule forward passes & 22.1\,s & 149.2\,s & GPU \\ mean-difference & 136.7\,s & 201.0\,s & CPU \\ classifier fitting & 311.6\,s & 451.2\,s & CPU \\ dump read & 29.4\,s & 142.4\,s & disk \\ \midrule SHIFT, total & \textbf{8.3\,min} & \textbf{15.8\,min} & \\ \midrule EraseAnything & \multicolumn{2}{c}{48 min} & GPU \\ \bottomrule \end{tabular} \end{table}



\paragraph{Storage.} Table~\ref{tab:storage} lists what an
erasure concept occupies at inference. The steering field does not grow with $n$, since it is a mean over pairs, so a concept ships as a few hundred
megabytes regardless of how many contrastive pairs were used to estimate it. The classifier ensemble, not the direction itself, is the dominant term.

\begin{table}[h]
\centering\footnotesize
\setlength{\tabcolsep}{5pt}
\caption{Storage per concept ($n = 135$).}
\label{tab:storage}
\begin{tabular}{@{}lr@{}}
\toprule
artifact & size \\
\midrule
steering field    & 57\,MB \\
classifier models & 182\,MB \\
text vectors      & 4\,MB \\
classifier scores & ${<}1$\,MB \\
\midrule
total             & \textbf{243\,MB} \\
\bottomrule
\end{tabular}
\end{table}

\subsection{Linear separability at the steering site}
\label{app:svm}

The classifier factor in Eq.~\ref{eq:alpha-diff} relies on the presence of the target concept being linearly readable from post-block text activations. We test this assumption using diagnostic linear classifiers. For each concept, block $\ell$, and denoising step $\tau$, we fit a linear SVM on token-mean activations and evaluate it on a held-out split grouped by contrastive pair: the positive and
negative members of each pair are always assigned to the same split.

\paragraph{At the first denoising step.}

Table~\ref{tab:svm_val} reports validation ROC-AUC at $\tau=0$, the step from which SHIFT derives its inference-time steering quantities. Linear separability
is consistently high: the mean across all concepts and blocks is $0.951$, and the lowest ROC-AUC at any individual site is $0.771$. Snoopy remains well
separated at every evaluated block.

\begin{table}[h]
\centering\scriptsize
\setlength{\tabcolsep}{2.6pt}
\caption{Validation ROC-AUC of the linear classifier at $\tau = 0$, per double-stream block. Chance is $0.5$; $n = 108$ held-out samples per site ($n = 16$ for Snoopy).}
\label{tab:svm_val}
\begin{tabular}{@{}lrrrrrrrr@{}}
\toprule
$\ell$ & hate & harass. & viol. & self-h. & sex. & shock. & illeg. & Snoopy \\
\midrule
0 & .949 & .986 & .897 & .888 & .983 & .874 & .880 & 1.00 \\
1 & .852 & .951 & .854 & .893 & .918 & .788 & .827 & 1.00 \\
2 & .820 & .928 & .833 & .877 & .893 & .771 & .868 & 1.00 \\
3 & .894 & .946 & .843 & .950 & .906 & .799 & .937 & 1.00 \\
4 & .962 & .973 & .855 & .972 & .928 & .872 & \textbf{.984} & 1.00 \\
5 & .993 & .993 & .898 & .981 & .969 & .951 & .983 & 1.00 \\
6 & \textbf{.999} & \textbf{.998} & .935 & .982 & \textbf{.993} & \textbf{.982} & \textbf{.984} & 1.00 \\
7 & .984 & .994 & .951 & \textbf{.995} & \textbf{.993} & .944 & .965 & 1.00 \\
8 & .969 & .977 & .941 & .989 & .983 & .917 & .962 & 1.00 \\
9 & .936 & .975 & .942 & \textbf{.995} & .977 & .911 & .946 & 1.00 \\
10 & .912 & .947 & .921 & .988 & .971 & .902 & .924 & 1.00 \\
11 & .958 & .961 & .941 & .988 & .974 & .946 & .953 & 1.00 \\
12 & .957 & .948 & .933 & .984 & .969 & .958 & .959 & 1.00 \\
13 & .971 & .954 & .944 & .983 & .973 & .968 & .972 & 1.00 \\
14 & .976 & .974 & .970 & .989 & .982 & .973 & .974 & 1.00 \\
15 & .979 & .976 & .977 & .990 & .983 & .970 & .975 & 1.00 \\
16 & .977 & .983 & .987 & \textbf{.995} & .987 & .971 & .975 & 1.00 \\
17 & .985 & .991 & \textbf{.989} & .984 & .991 & .968 & .976 & 1.00 \\
18 & .794 & .903 & .886 & .852 & .920 & .820 & .867 & 1.00 \\
\midrule
mean & .940 & .966 & .921 & .962 & .963 & .910 & .943 & 1.00 \\
\bottomrule
\end{tabular}
\end{table}

\paragraph{Across denoising steps.}

\begin{figure}[h]
  \centering
  \includegraphics[width=\linewidth]{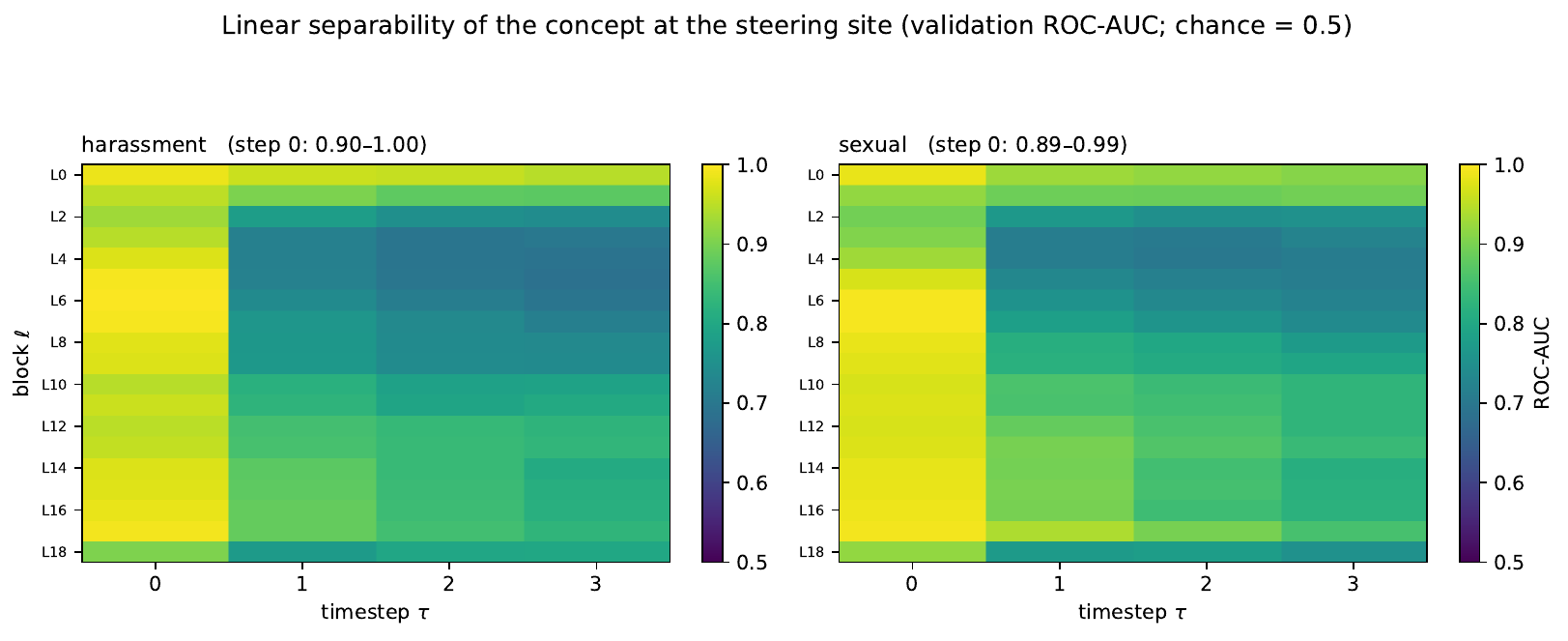}
  \caption{Validation ROC-AUC of the linear classifier over all blocks and denoising steps for two representative concepts. }
  \label{fig:svm_val}
\end{figure}

Fig.~\ref{fig:svm_val} shows the complete block--step grid for two representative concepts. Separability decreases at later steps, consistent with the attribution analysis of Mazur et al.~\cite{mazur2026diffract}, who find that the share of attribution carried by text-stream features in Flux.1[schnell] decreases from $89.9\%$ at step $0$ to $5.4\%$ at step $3$. The early-step regime of strong linear readability also coincides with the interval over which the steering direction remains stable (App.~\ref{app:geometry}) and the intervention remains
effective (Table~\ref{tab:block_ablation_wide}). Together, these complementary measurements support the use of early text-stream activations for steering.

\paragraph{Choosing the strength.}
The nominal values of $\gamma_{\mathrm{diff}}$ differ across configurations for two reasons that do not involve per-concept tuning. First, $\gamma_{\mathrm{diff}}$ is only a base scale that is always multiplied by the gate; a nominal value of $225$ therefore corresponds to a median effective strength between $0.69$ and $10.90$ depending on the concept (App.~\ref{app:gate}). The gate, rather than the operator, adapts the intervention to each prompt. Consequently, even at a single fixed $\gamma_{\mathrm{diff}}$ the applied strength already varies by an order of magnitude across the seven inappropriate categories, which is why one value transfers without re-tuning.
Second, the operating regime of the gate itself depends on whether the target is a concrete named object or an abstract concept. When the target is a named object, the prompt either contains the name or it does not, and the gate effectively becomes a binary switch: on the Snoopy protocol it saturates on $100\%$ of target prompts (effective strength $27.95$), while on the five preserved objects it remains fully closed on $47\%$ of prompts and saturates on only $1.5\%$. For an abstract concept the gate operates almost entirely below saturation: on I2P only $4.2\%$ of nudity prompts reach $\delta$ and the median gate value is $0.038$, so the same nominal $\gamma_{\mathrm{diff}}$ yields a median effective strength an order of magnitude lower.

\section{Additional Flux.1 quality results}

\begin{figure}[h!]
    \centering
    \includegraphics[width=1\linewidth]{figures/coco.pdf}
    \caption{Examples of image generations on the COCO dataset using Flux.1[dev], Flux.1[schnell], and attn- and block-based steered Flux.1[dev]*, Flux.1[dev] and Flux.1[schnell]  generations, where * denotes steering Flux.1[dev] with the activation vector derived from Flux.1[schnell].}
    \label{fig:coco}
\end{figure}

\subsection{Flux.1[dev] nudity erase}
\begin{figure}[h!]
    \centering
    \includegraphics[width=1\linewidth]{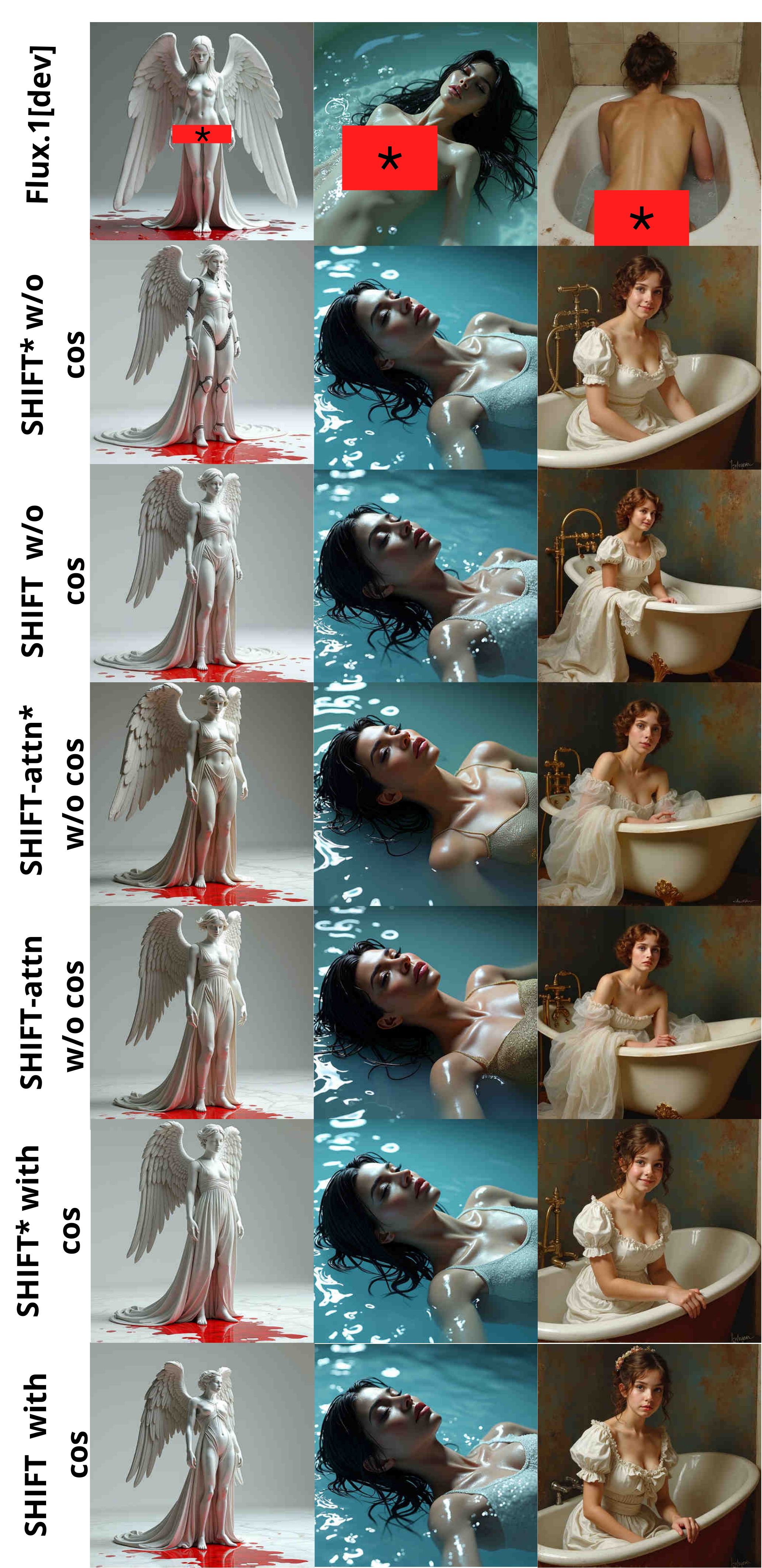}
    \caption{Qualitative results of nudity concept erasure on Flux.1[dev]. Comparison between the baseline (no steering), our main residual-block method (SHIFT), and the attention-based variant (SHIFT-attn). * denotes that we use steering vector from Flux.1[schnell] activations to steer Flux.1[dev]}
    \label{fig:nud_all_dev}
\end{figure}

In addition to the main experiments on nudity concept erasure, we evaluate the image generation quality of our method on the COCO dataset using 5,000 prompts. We report FID and CLIP scores for both attention-based (SHIFT-attn) and residual-block (SHIFT) steering. The results are presented in Table~\ref{tab:dev_nudity_coco}, qualitative results are shown in Figure~\ref{fig:nud_all_dev}.
The steering strength introduces a clear trade-off between overall image quality and erasure effectiveness. Notably, residual-block steering degrades image quality (FID) significantly less than attention-based steering. We additionally provide qualitative examples of images generated with and without steering for nudity erasure using COCO prompts in Figure~\ref{fig:coco}.

\begin{table}[t]
\centering
\caption{Quantitative evaluation of Flux.1[dev] steering for nudity concept erasure and overall image quality metrics (FID and CLIP) on the COCO dataset using 5k prompts.}
\label{tab:dev_nudity_coco}
\scriptsize
\setlength{\tabcolsep}{3pt}
\resizebox{\linewidth}{!}{%
\begin{tabular}{llrrrrrr}
\toprule
\textbf{Method} & \textbf{Strength} & \multicolumn{4}{c}{\textbf{Detected Nudity (Quantity)}} & \multicolumn{2}{c}{\textbf{MS-COCO 5K}} \\
\cmidrule(lr){3-6}\cmidrule(lr){7-8}
 &  & \textbf{Common} & \textbf{Female} & \textbf{Male} & \textbf{Total$\downarrow$} & \textbf{FID$\downarrow$} & \textbf{CLIP$\uparrow$} \\
\midrule
SHIFT-attn (schnell acts) & 250 & 106 & 31 & 15 & 152 & 37.90 & \underline{30.69} \\
SHIFT-attn (dev acts) & 250 & 155 & 43 & 17 & 215 & 38.50 & \underline{30.69} \\
\midrule
SHIFT (dev acts) & 30 & \textbf{62} & \textbf{11} & \textbf{6} & \textbf{79} & 37.30 & 29.70 \\
SHIFT (schnell acts) & 30 & \textbf{62} & \underline{14} & \underline{7} & \underline{83} & 37.00 & 30.00 \\
\midrule
SHIFT with cos (schnell acts) & 250 & 117 & 25 & 14 & 156 & \textbf{35.00} & \underline{30.69} \\
SHIFT with cos (schnell acts) & 280 & \underline{96} & 18 & 12 & 126 & \textbf{35.00} & 30.68 \\
\midrule
Flux.1[dev] & -- & 406 & 161 & 38 & 605 & \underline{35.83} & \textbf{30.91} \\
\bottomrule
\end{tabular}%
}
\end{table}

\begin{figure}[h!]
    \centering
    \includegraphics[width=1\linewidth]{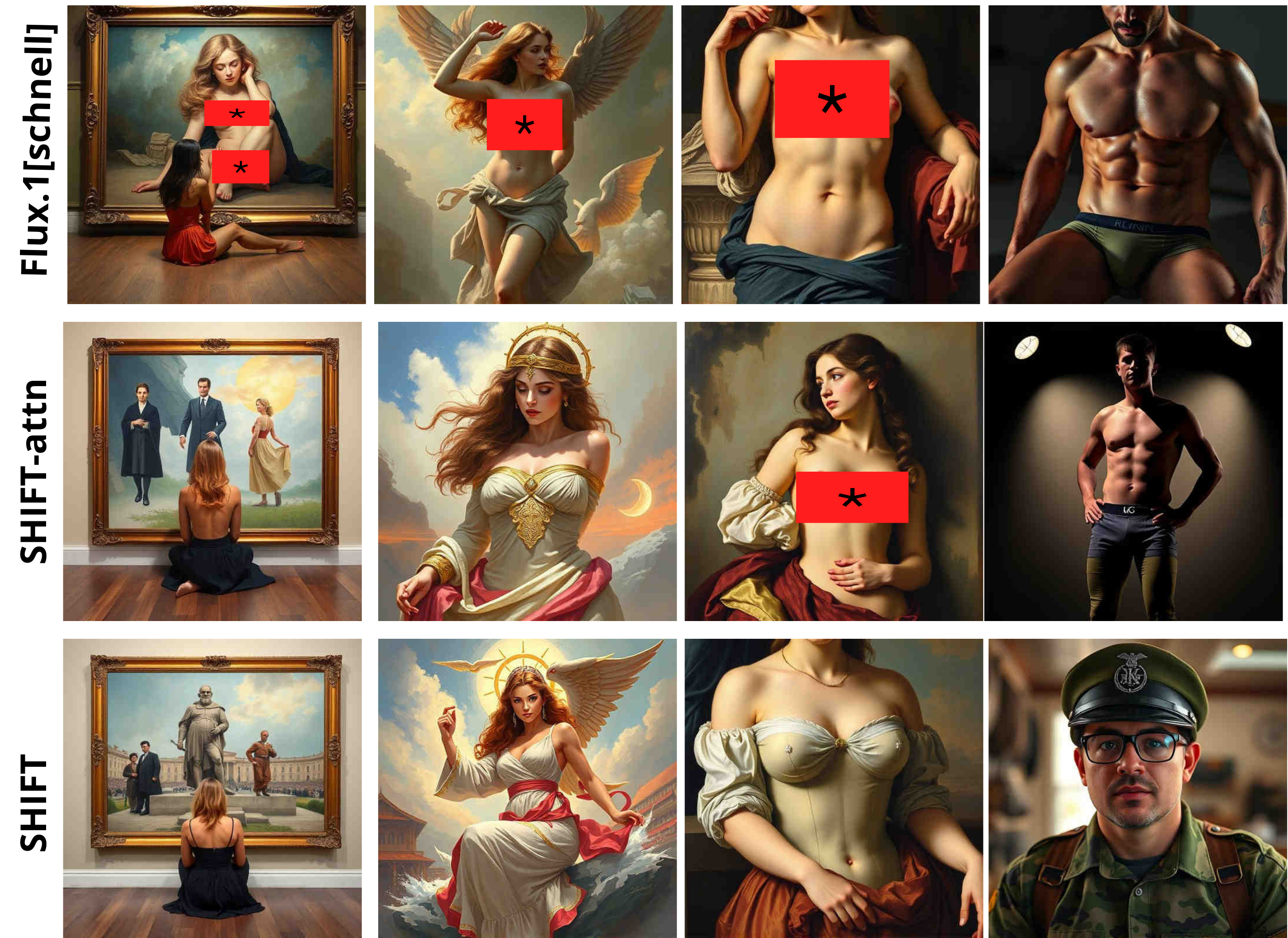}
    \caption{Qualitative results of nudity concept erasure on Flux.1[schnell]. Comparison between the baseline (no steering), our main residual-block method (SHIFT), and the attention-based variant (SHIFT-attn).}
    \label{fig:nud_all_schnell}
\end{figure}

\subsection{Multi-concept erasure on Flux.1[dev]}
We additionally evaluate joint multi-concept erasure on the non-distilled
Flux.1[dev] model, following the same protocol as for Flux.1[schnell]: the seven
inappropriate I2P categories are erased simultaneously with a single combined
steering field, and inappropriate content is detected with the Q16
classifier. We report results for steering vectors computed on
Flux.1[dev]'s own activations as well as for vectors transferred from the
distilled Flux.1[schnell], assessing whether the cross-distillation transfer
observed for single-concept erasure also holds in the multi-concept setting.
Quantitative results are presented in Table~\ref{tab:multi-concept-dev}. As in
the single-concept case, SHIFT remains effective on the larger model, and the
clipped variant gives a consistent improvement over the unclipped one --- with
transferred vectors performing comparably to those computed natively,
confirming that distillation preserves the semantic geometry exploited by our
steering even when several concepts are removed at once.
\begin{table}[t]
\centering
\caption{Multi-concept erasure on the I2P~\cite{schramowski2023safe} dataset for Flux.1[dev].
Detection is done by the Q16~\cite{schramowski2022can} classifier. * denotes steering vectors
transferred from Flux.1[schnell]. The best results are highlighted in
\textbf{bold}.}
\label{tab:multi-concept-dev}
\resizebox{\linewidth}{!}{%
\begin{tabular}{l|ccc}
\toprule
\multirow{2}{*}{Class name} & \multicolumn{3}{c}{Inappropriate proportion (\%) ($\downarrow$)}\\
\cmidrule(l){2-4}
 & Flux.1[dev] & EraseAnything & SHIFT*\\
\midrule
Hate             & 33.33 & 24.68 & \textbf{9.69}\\
Harassment       & 28.40 & 23.79 & \textbf{10.35}\\
Violence         & 32.80 & 23.41 & \textbf{9.14}\\
Self-harm        & 41.45 & 29.71 & \textbf{18.92}\\
Sexual           & 49.62 & 41.03 & \textbf{23.70}\\
Shocking         & 47.43 & 33.53 & \textbf{18.40}\\
Illegal activity & 28.20 & 21.18 & \textbf{8.84}\\
\midrule
Overall          & 38.42 & 29.19 & \textbf{15.37}\\
\midrule
FID$\downarrow$ (COCO 5K)  & 35.83 & 36.30 & \textbf{33.76}\\
CLIP$\uparrow$ (COCO 5K)   & \textbf{30.91} & 30.46 & 28.58\\
\bottomrule
\end{tabular}}
\end{table}

\subsection{Concrete object steering for Flux.1[dev]}\label{sec:_app_dev_snoopy}
We evaluate Snoopy concept removal on the Flux.1[dev] model. Following~\cite{lyu2024one}, we use 80 original prompts and 9 different random seeds per prompt for validation. We additionally test our method on Snoopy erasure while preserving related concepts such as Mickey, SpongeBob, Pikachu, dog, and legislator. For both attention-based (SHIFT-attn) and residual-block steering, we assess generation quality using CLIP scores (between target prompts and generated images) and FID. While the ESD baseline fails to erase the target concept, our method achieves significantly better erasure performance, as shown in Table~\ref{tab:object-erasure_dev}.

\begin{table}[t]
  \centering
  \caption{Quantitative evaluation of concrete object erasure for Flux.1[dev] model.}
  \label{tab:object-erasure_dev}
  \setlength{\tabcolsep}{3pt}
  \resizebox{\linewidth}{!}{%
    \begin{tabular}{lcccccccccccc}
      \toprule
      & \multicolumn{2}{c}{Snoopy} & \multicolumn{2}{c}{Mickey} & \multicolumn{2}{c}{SpongeBob} & \multicolumn{2}{c}{Pikachu} & \multicolumn{2}{c}{Dog} & \multicolumn{2}{c}{Legislator} \\
      \cmidrule(lr){2-3}\cmidrule(lr){4-5}\cmidrule(lr){6-7}\cmidrule(lr){8-9}\cmidrule(lr){10-11}\cmidrule(lr){12-13}
      Method & CS$\downarrow$ & FID & CS$\uparrow$ & FID$\downarrow$ & CS$\uparrow$ & FID$\downarrow$ & CS$\uparrow$ & FID$\downarrow$ & CS$\uparrow$ & FID$\downarrow$ & CS$\uparrow$ & FID$\downarrow$ \\
      \midrule
      ESD & 26.77 & 15.44 & 26.18 & 6.56 & 27.09 & 9.51 & 26.81 & 7.87 & 24.16 & 8.76 & 20.34 & 8.52 \\
      SHIFT-attn (dev acts) & 11.98 & 181.00 & 22.70 & 88.70 & 26.13 & 67.19 & 25.14 & 51.12 & 22.98 & 56.00 & 17.90 & 72.95 \\
      \midrule
      SHIFT w/o cos (dev acts) & 16.42 & 158.91 & 26.44 & 32.68 & 26.95 & 36.82 & 26.49 & 27.46 & 24.22 & 37.05 & 19.87 & 32.45 \\
      SHIFT w/o cos (schnell acts) & 17.01 & 158.14 & 26.44 & 31.88 & 27.06 & 36.26& 26.55 & 26.86 & 24.29 & 37.02 & 20.01 & 31.64 \\
      SHIFT with cos (schnell acts) & 13.78 & 148.72 & 26.35 & 31.88 & 26.78 & 36.26& 26.65 & 26.86 & 23.6 & 37.02 & 20.71 & 31.64 \\
      \midrule
      Flux.1[dev] & 27.43 & -- & 26.20 & -- & 27.14 & -- & 26.90 & -- & 24.19 & -- & 20.51 & -- \\
      \bottomrule
    \end{tabular}%
  }
\end{table}

\section{Style erasure}
Additionally, we tested our method for style erasure. First, we assessed the ability to remove the style of a famous artist like Van Gogh while preserving similar artists (Picasso, Rembrandt, Warhol, Caravaggio). For the second task, we evaluated the removal of a modern artist's style like McKernan, while preserving Kinkade, Edlin, Eng, and Ajin: Demi-Human.

Artist and style concepts are densely packed in embedding space, so a target style is often weakly correlated with many unrelated prompts. To prevent leakage onto these near-neighbour styles, we add a hard threshold below which no
steering is applied before the usual clip. Given the raw cosine $\hat c = \cos\langle e_\text{init}, e_\text{target}\rangle$, we set
\begin{equation}
  c \;=\;
  \begin{cases}
    0, & \hat c < \delta_1,\\[2pt]
    \operatorname{clip}(\hat c,\, 0,\, \delta), & \hat c \ge \delta_1,
  \end{cases}
  \label{eq:style-gate}
\end{equation}
where $\delta_1$ is a style-specific cut-off. Zeroing all sub-threshold
coefficients excises prompts that are only marginally related to the target
style, while the subsequent clip to $[0,\delta]$ bounds the strength on the
remaining on-style prompts. In practice this sharpens the boundary between the
erased style and visually adjacent ones, removing the target while leaving other
artists' styles intact. For style erasure we set the pooled text-encoder steering strength to $5$ and the DiT block steering strength to $50$, with the style-specific gate restricted to $c \in [0.2, 0.5]$ (i.e., $\delta_1 = 0.2$, $\delta = 0.5$).

\begin{figure}[h!]
    \centering
    \includegraphics[width=1\linewidth]{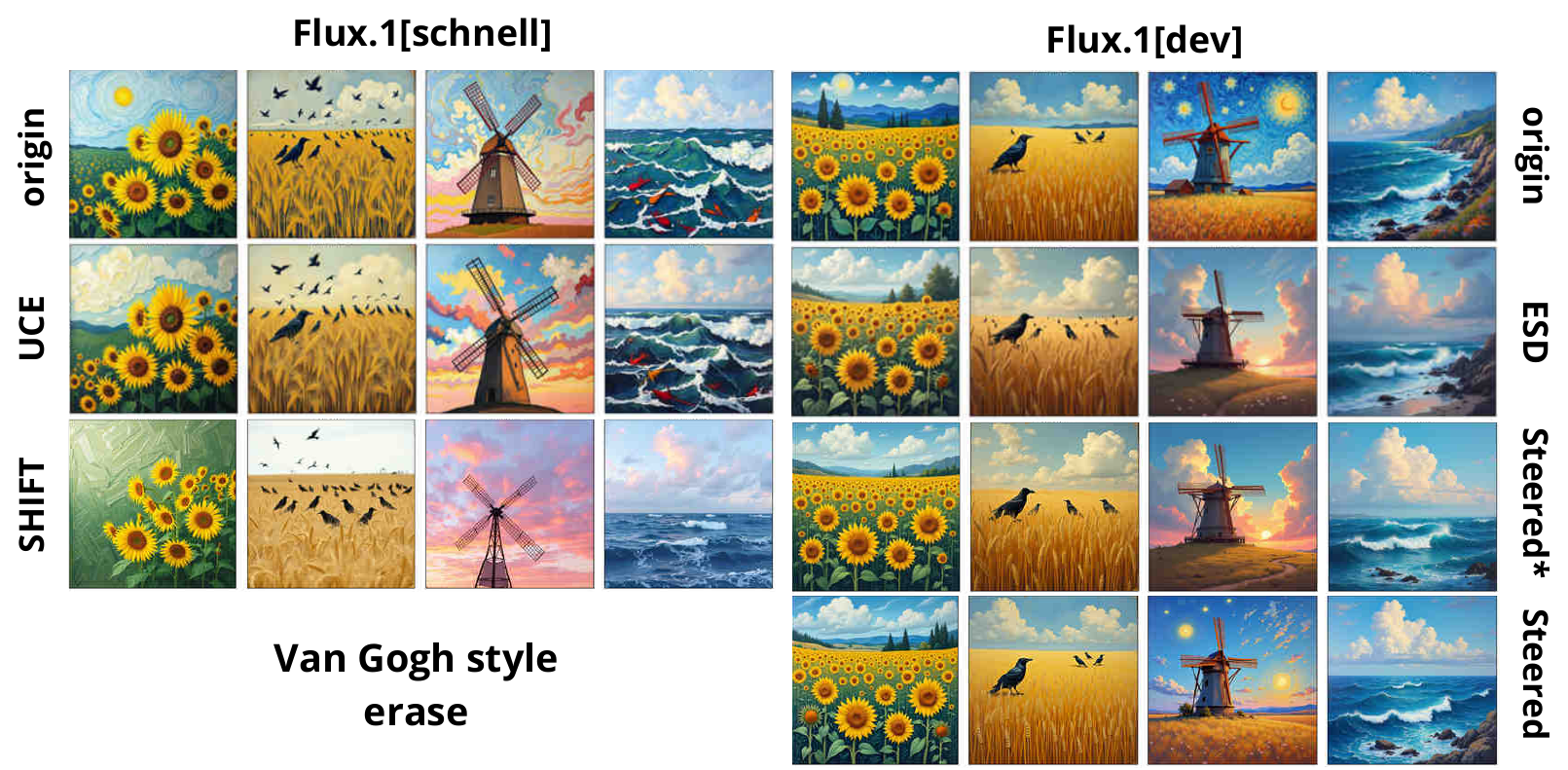}
    \caption{Qualitative results for Van Gogh style erasure. First row: original Flux.1[dev] and Flux.1[schnell] generations. Second row: UCE and ESD competitors. Last rows: our steered generations, where * denotes steering Flux.1[dev] with the activation vector derived from Flux.1[schnell].}
    \label{fig:artists_big}
\end{figure}

\begin{figure}[h!]
    \centering
    \includegraphics[width=1\linewidth]{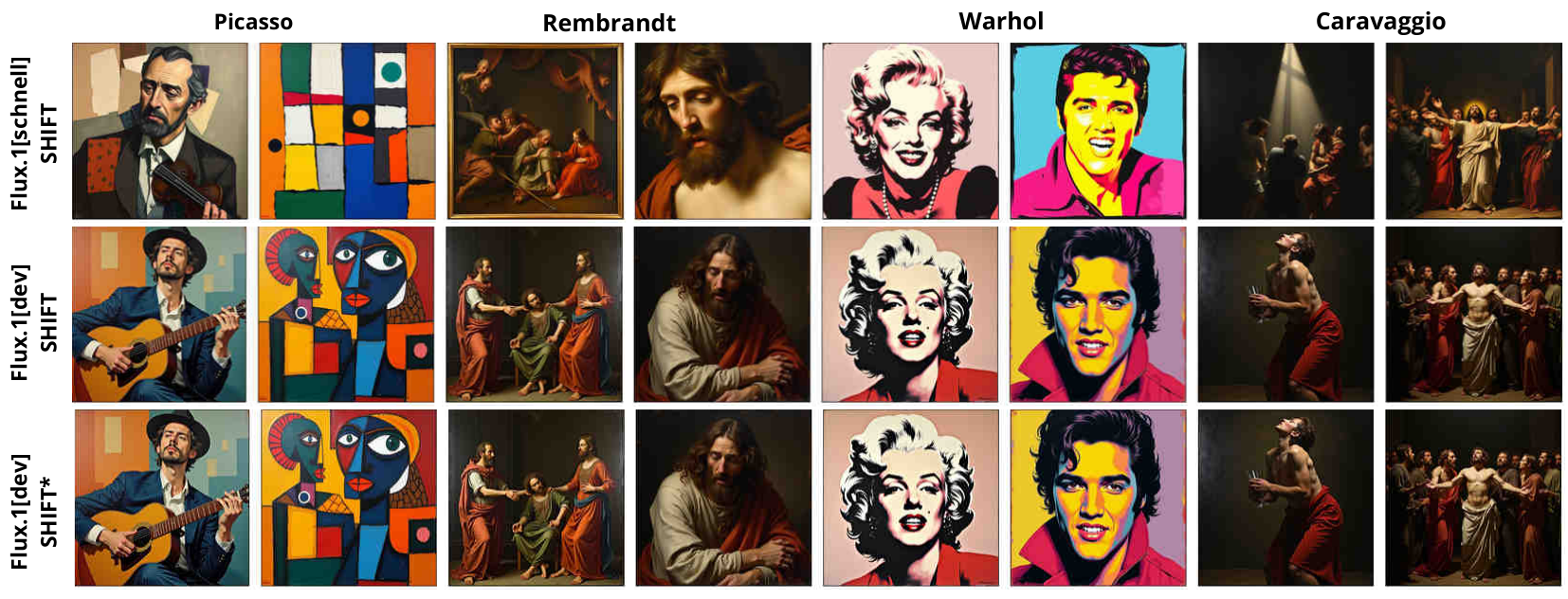}
    \caption{Qualitative results demonstrating preservation of other artists' styles (Picasso, Rembrandt, Warhol, Caravaggio) while erasing Van Gogh's style using our steered model (SHIFT).}
    \label{fig:artists_big_other}
\end{figure}

For validation, we used GPT-4o to identify artists in generated images as an evaluation metric. Questions to GPT-4o were formulated as ``Is this picture in $s$'s style? Just tell me Yes or No.'', where $s$ represents the style. We estimated the accuracy of detected artists. We compared our model with ESD (for Flux.1[dev]) and UCE (for Flux.1[schnell] model) baselines. The quantitative results are presented in Table~\ref{tab:style_erase_accuracy}, and qualitative
results are presented in Fig.~\ref{fig:artists_big},~\ref{fig:artists_big_other} for Van Gogh erasure.

\begin{table}[t]
    \centering
    \small
    \setlength{\tabcolsep}{6pt}
    \renewcommand{\arraystretch}{1.1}
    \caption{Style-erasure accuracy on old and modern artist groups. $\mathrm{Acc}_{E}$ denotes accuracy on erased-target prompts, and $\mathrm{Acc}_{U}$ denotes accuracy on unaffected prompts.}
    \label{tab:style_erase_accuracy}
    \begin{tabular}{lcccc}
        \toprule
        \textbf{Method} & \multicolumn{2}{c}{\textbf{Old artists}} & \multicolumn{2}{c}{\textbf{Modern artists}} \\
        \cmidrule(lr){2-3} \cmidrule(lr){4-5}
        & $\mathbf{Acc_{E}\downarrow}$ & $\mathbf{Acc_{U}\uparrow}$ & $\mathbf{Acc_{E}\downarrow}$ & $\mathbf{Acc_{U}\uparrow}$ \\
        \midrule
        ESD & 20.00\% & 63.75\% &  68.33\% & 67.08\% \\
        SHIFT [dev] &   \textbf{10.00\%} & \textbf{76.25\%} & \textbf{31.67\%} & \textbf{68.33\%} \\
        SHIFT [schnell]&  20.00\% & \textbf{76.25\%} & \textbf{31.67\%} & 67.50\% \\
        \midrule
        Flux.1 [dev] & 68.33\% & 79.58\% & 88.33\% & 81.25\%  \\
        \midrule
        UCE & 23.33\% & 75.41\% & 61.67\% & 64.16\% \\
        SHIFT  & \textbf{20.00\%} & \textbf{81.25\%} & \textbf{30.00\%} & \textbf{75.25\%} \\
        \midrule
        Flux.1 [schnell] & 80.00\% & 84.59\% & 88.33\% & 85.84 \% \\
       
        \bottomrule
    \end{tabular}
\end{table}

\section{Generalization to SD3.5.}
To verify that SHIFT is not specific to Flux, we apply it to another MM-DiT
model, SD3.5, on the concrete object erasure task. Following the same protocol as
for Flux.1[schnell], we erase the Snoopy concept while preserving five related
concepts (Mickey, SpongeBob, Pikachu, dog, and legislator), using pooled text-encoder strength $5$, DiT block steering strength $50$, and cosine gate bound $\delta = 0.3$. As shown in
Table~\ref{tab:sd35_snoopy}, SHIFT sharply reduces the CLIP score of the target
concept (Snoopy: $29.19\to18.80$) while leaving the preserved concepts almost
unchanged, with their CLIP scores deviating only marginally from the original
model. This confirms that the residual-structure steering underlying SHIFT
transfers across MM-DiT architectures and is not tied to a specific backbone.

\begin{table}[h!]
  \centering
  \caption{Quantitative evaluation of concrete object erasure (Snoopy) for the SD3.5 model.}
  \label{tab:sd35_snoopy}
  \setlength{\tabcolsep}{3pt}
  \resizebox{\linewidth}{!}{%
    \begin{tabular}{lcccccccccccc}
      \toprule
      & \multicolumn{2}{c}{Snoopy} & \multicolumn{2}{c}{Mickey} & \multicolumn{2}{c}{SpongeBob} & \multicolumn{2}{c}{Pikachu} & \multicolumn{2}{c}{Dog} & \multicolumn{2}{c}{Legislator} \\
      \cmidrule(lr){2-3}\cmidrule(lr){4-5}\cmidrule(lr){6-7}\cmidrule(lr){8-9}\cmidrule(lr){10-11}\cmidrule(lr){12-13}
      Method & CS$\downarrow$ & FID & CS$\uparrow$ & FID$\downarrow$ & CS$\uparrow$ & FID$\downarrow$ & CS$\uparrow$ & FID$\downarrow$ & CS$\uparrow$ & FID$\downarrow$ & CS$\uparrow$ & FID$\downarrow$ \\
      \midrule
      SHIFT & \textbf{18.80} & 173.49 & 26.66 & 37.04 & 28.79 & 20.20 & 27.59 & 12.84 & 24.57 & 18.19 & 22.51 & 1.58 \\
      \midrule
      SD3.5 & 29.19 & -- & 26.76 & -- & 28.78 & -- & 27.56 & -- & 24.62 & -- & 22.50 & -- \\
      \bottomrule
    \end{tabular}%
  }
\end{table}

\begin{table}[h!]
\centering
\caption{(Left) Quantity of explicit content detected using the NudeNet detector on the I2P benchmark for SD3.5. (Right) Comparison of FID and CLIP on MS-COCO 5K sampled images.}
\label{tab:detected_nudity_sd35}
\footnotesize
\setlength{\tabcolsep}{2.0pt}
\renewcommand{\arraystretch}{1.0}
\resizebox{\columnwidth}{!}{%
\begin{tabular}{lcccccc}
\toprule
\textbf{Method} & \multicolumn{4}{c}{\textbf{Detected Nudity (Quantity)}} & \multicolumn{2}{c}{\textbf{MS-COCO 5K}} \\
\cmidrule(lr){2-5}\cmidrule(lr){6-7}
 & \textbf{Common} & \textbf{Female} & \textbf{Male} & \textbf{Total$\downarrow$} & \textbf{FID$\downarrow$} & \textbf{CLIP$\uparrow$} \\
\midrule
SHIFT & \textbf{17} & \textbf{6} & \textbf{6} & \textbf{29} & 27.60 & 31.74 \\
\midrule
SD3.5 & 167 & 81 & 19 & 267 & 28.29 & 31.76 \\
\bottomrule
\end{tabular}%
}
\end{table}

\section{Additional ablations}
\subsection{Nudity strength steering}
We further provide ablations on the effect of steering strength for nudity erasure. These results, evaluated on 5,000 MS-COCO prompts using FID and CLIP metrics, are presented in Table~\ref{tab:schnell_nudity_strength}. The analysis confirms that steering acts as a tunable trade-off between image quality preservation and effective concept erasure.

\begin{table}[t]
\centering
\caption{Quantitative evaluation of Flux.1[schnell] steering for nudity concept erasure and overall image quality metrics (FID and CLIP) on the COCO dataset using 5k prompts for different strength.}
\label{tab:schnell_nudity_strength}
\scriptsize
\setlength{\tabcolsep}{3pt}
\resizebox{\linewidth}{!}{%
\begin{tabular}{llrrrrrr}
\toprule
\textbf{Method} & \textbf{Strength} & \multicolumn{4}{c}{\textbf{Detected Nudity (Quantity)}} & \multicolumn{2}{c}{\textbf{MS-COCO 5K}} \\
\cmidrule(lr){3-6}\cmidrule(lr){7-8}
 &  & \textbf{Common} & \textbf{Female} & \textbf{Male} & \textbf{Total$\downarrow$} & \textbf{FID$\downarrow$} & \textbf{CLIP$\uparrow$} \\
\midrule
SHIFT & 25 & 76 & 30 & 6 & 112 & 32.16 & 31.42 \\
SHIFT & 30 & 61 & 20 & 4 & 85 & 32.3 & 31.18 \\
SHIFT  & 35 & 42 & 8 & 2 & 52 & 32.5 & 30.84 \\
\midrule
Flux.1[schnell] & -- &  412 & 190 & 10 & 612 & 31.63 & 31.59 \\
\bottomrule
\end{tabular}%
}
\end{table}

\subsection{Injection blocks and temporal dynamics}

\begin{figure}[h!]
    \centering
    \includegraphics[width=1\linewidth]{figures/blocks_abl.pdf}
    \caption{Ablation on steering applied at different blocks of the DiT backbone for concrete object (Snoopy) erasure. * denotes that we use different steering vectors for different steps.}
    \label{fig:block_abl}
\end{figure}

We ablate the influence of steering across different blocks of the DiT backbone in Table~\ref{tab:block_ablation_wide} and Figure~\ref{fig:block_abl}. We also evaluate temporal dynamics and compare a single shared steering vector with timestep-specific steering vectors. Our results show that steering is most effective when applied during the early stages of the diffusion trajectory, while steering in the latter half yields insufficient concept suppression. Importantly, using a single shared steering vector achieves comparable erasure performance to timestep-specific vectors without degrading image quality.

\begin{table*}[t]
  \centering
  \small 
  \caption{Ablation of steering across block ranges and timestep schedules.}
  \label{tab:block_ablation_wide}
  \setlength{\tabcolsep}{6pt} 
  \begin{tabular}{ll cc cc cc cc cc cc}
    \toprule
    & & \multicolumn{2}{c}{Legislator} & \multicolumn{2}{c}{Mickey} & \multicolumn{2}{c}{Pikachu} & \multicolumn{2}{c}{SpongeBob} & \multicolumn{2}{c}{Dog} & \multicolumn{2}{c}{Snoopy} \\
    \cmidrule(lr){3-4}\cmidrule(lr){5-6}\cmidrule(lr){7-8}\cmidrule(lr){9-10}\cmidrule(lr){11-12}\cmidrule(lr){13-14}
    Blocks & Steps & CS$\uparrow$ & FID$\downarrow$ & CS$\uparrow$ & FID$\downarrow$ & CS$\uparrow$ & FID$\downarrow$ & CS$\uparrow$ & FID$\downarrow$ & CS$\uparrow$ & FID$\downarrow$ & CS$\downarrow$ & FID$\uparrow$ \\
    \midrule
    0 & 0 & 21.9 & -- & 26.7 & -- & 27.2 & -- & 27.9 & -- & 24.6 & -- & 28.0 & -- \\
    \midrule
    0--6 & all & 21.7 & 34.9 & 26.8 & 43.4 & 27.3 & 34.0 & 28.1 & 40.6 & 24.7 & 31.6 & 21.7 & 129.5 \\
    6--12 & all & 21.9 & 25.9 & 26.6 & 33.9 & 27.2 & 28.8 & 28.0 & 33.0 & 24.7 & 88.6 & 24.9 & 86.7 \\
    12--18 & all & 21.9 & 19.1 & 26.6 & 28.2 & 27.1 & 23.8 & 28.0 & 25.6 & 24.5 & 20.7 & 26.2 & 75.9 \\
    all & 0--1 & 21.6 & 40.69 & 26.5 & 51.4 & 27.2 & 40.0 & 28.0 & 47.6 & 24.7 & 36.9 & 18.8 & 150.9 \\
    all & 2--3 &21.9 & 12.0 & 26.6 & 25.2 & 27.2& 19.8 &28.0 & 18.7 & 24.6& 15.8 & 26.5& 70.6 \\
    all & all* &21.5 & 40.7 & 26.5& 51.6 & 27.3& 40.0 & 28.1& 47.9 & 24.6& 37.0 & 18.7 & 151.9 \\
    all & all & 21.5 & 40.5 & 26.5 & 51.7 & 27.2 & 40.1 & 27.8 & 46.6 & 24.6 & 37.2 & 18.7 & 151.4 \\
    \bottomrule
  \end{tabular}
\end{table*}

\subsection{Strength for concrete object erase ablation}
As shown in Figure~\ref{fig:strength_abl}, increasing the steering strength $  \alpha  $ leads to more effective erasure of the Snoopy concept. However, stronger steering also increases the risk of unintentionally degrading non-target concepts such as Mickey, SpongeBob, and legislator. This illustrates the inherent trade-off between erasure effectiveness and the preservation of unrelated concepts.

\subsection{Effect of cosine clipping}
\label{sec:clip-ablation}
\begin{figure}[!t]
    \centering
    \includegraphics[width=1\linewidth]{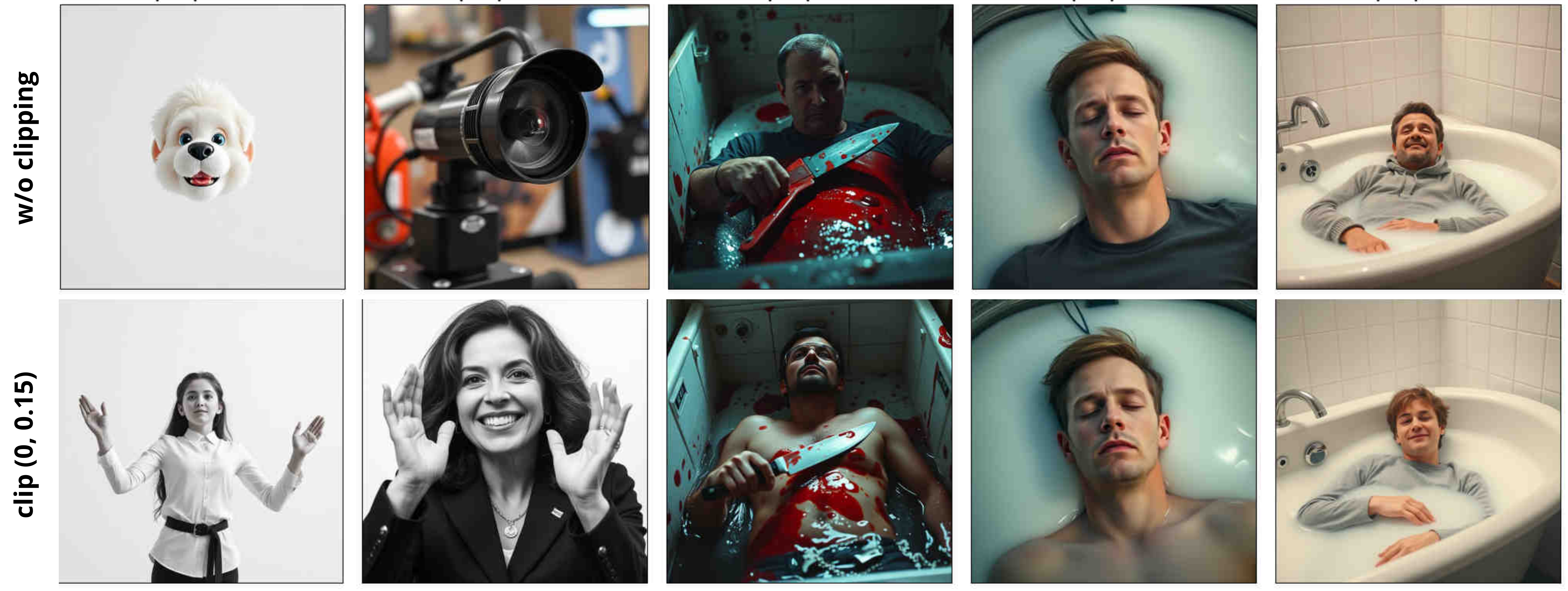}
    \caption{Effect of cosine clipping on nudity erasure for Flux.1[schnell].}
    \label{fig:clip-ablation-qual}
\end{figure}

We isolate the contribution of the cosine clipping introduced in
method section. The clip bounds the gate to $[0,\delta]$, capping the intervention strength on prompts that are strongly aligned with the target concept. Without this bound, the cosine gate grows unchecked on such prompts, and
the steering becomes so strong that the generation no longer follows the input
prompt: when erasing nudity, for instance, the model does not just  remove the unsafe content but can suppress the depicted person entirely, collapsing the image rather than redirecting it to a safe variant. Clipping prevents this failure mode by keeping the steering within a range that removes the target concept while preserving the rest of the prompt. A qualitative comparison is
shown in Fig.~\ref{fig:clip-ablation-qual}.

\section{Concept addition and switching}

\begin{figure}[!t]
    \centering
    \includegraphics[width=1\linewidth]{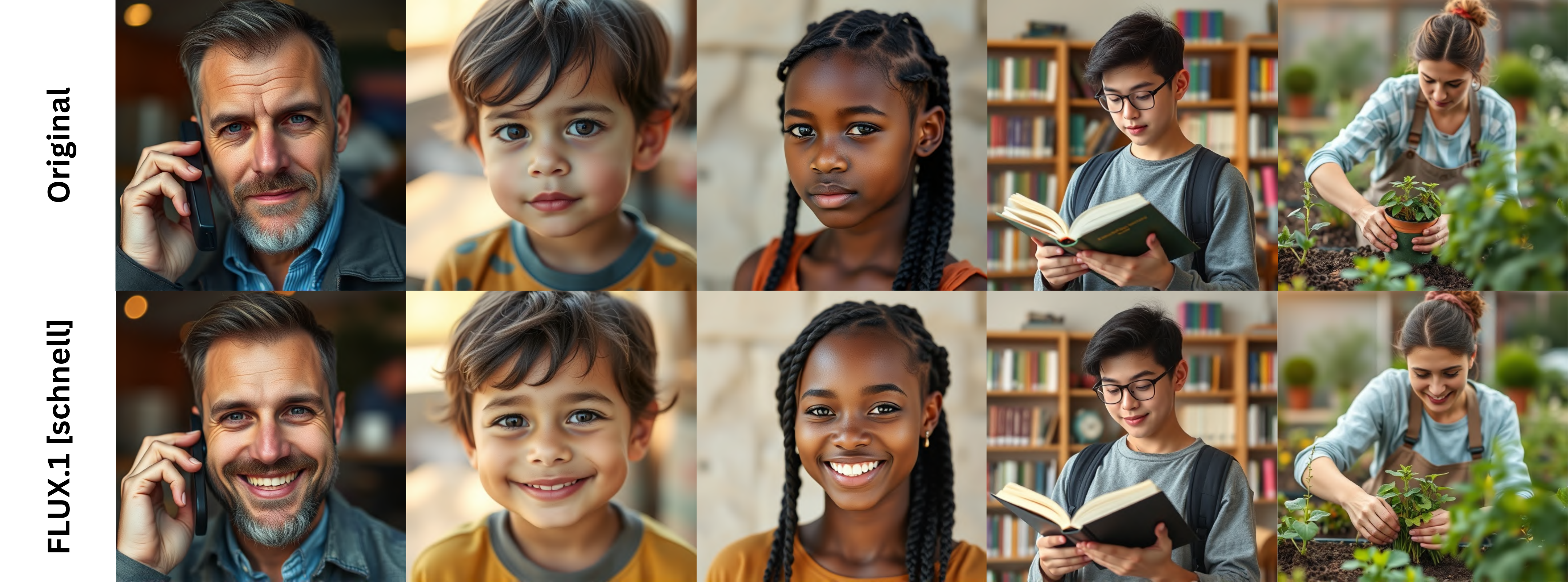}
    \caption{Steering to add smile}
    \label{fig:add_smile}
\end{figure}

\begin{figure}[!t]
    \centering
    \includegraphics[width=1\linewidth]{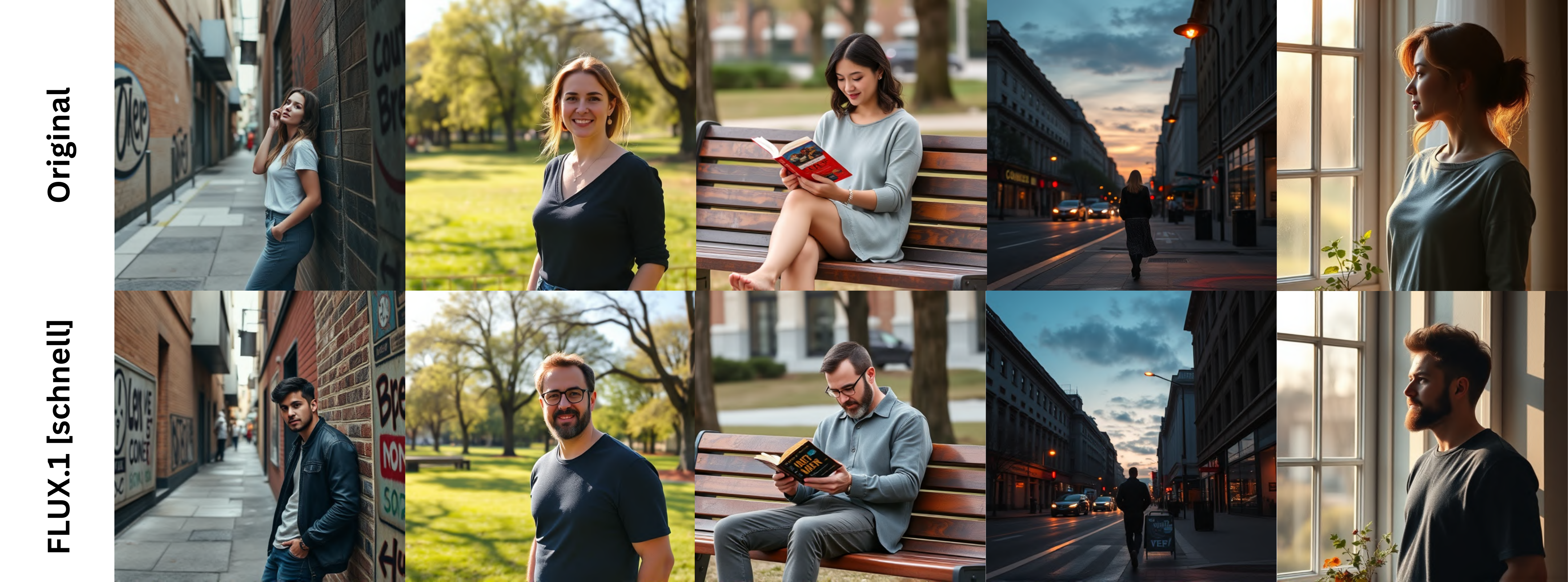}
    \caption{Steering to switch woman to man}
    \label{fig:switch_man}
\end{figure}

\begin{figure}[!t]
    \centering
    \includegraphics[width=1\linewidth]{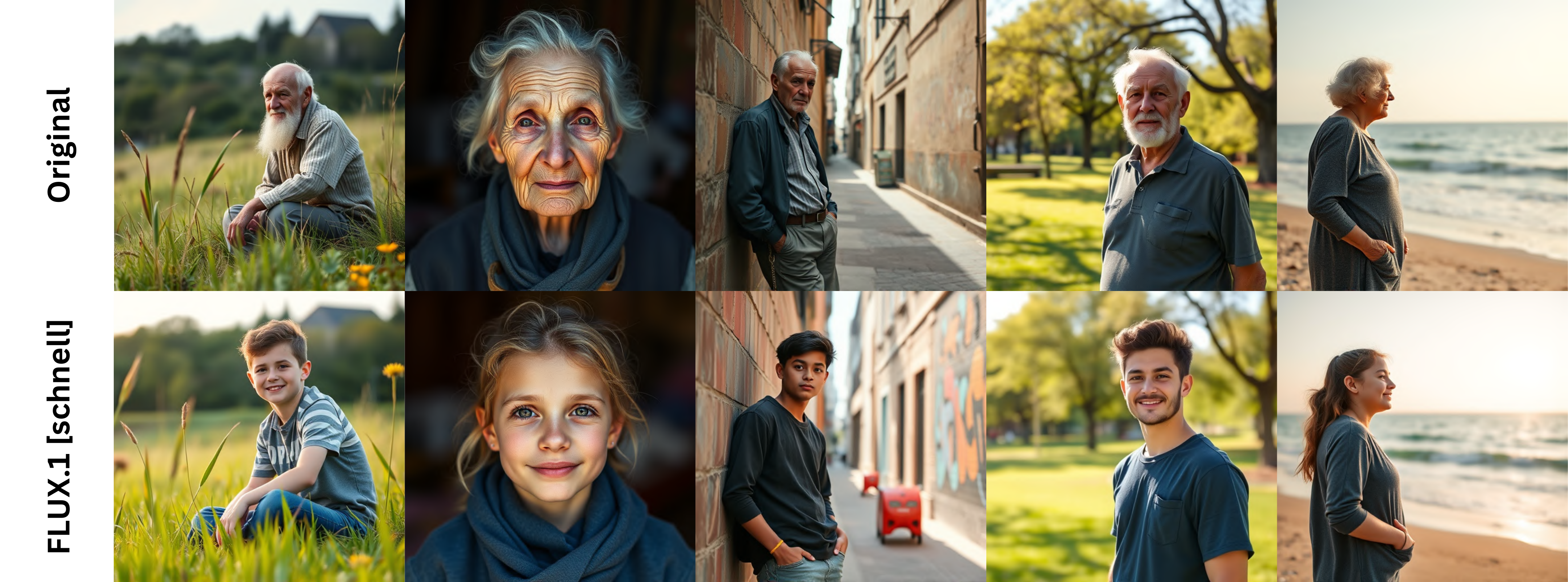}
    \caption{Steering to switch old to young}
    \label{fig:switch_young}
\end{figure}

We explore additional tasks in Figures~\ref{fig:add_smile}--\ref{fig:switch_young}, specifically addition of the concept (smile) and switching between concepts (woman to man, old to young), and showcase qualitative results.

\begin{figure}[h!]
    \centering
    \includegraphics[width=1\linewidth]{figures/strength_abl.pdf}
    \caption{Ablation on steering strength for Snoopy concept erasure on Flux.1[schnell]. From left to right: only text steering followed by increasing steering strength $\gamma$.}
    \label{fig:strength_abl}
\end{figure}

\section{Gate behaviour}
\label{app:gate}
\begin{figure*}[h!]
  \centering
  \includegraphics[width=\linewidth]{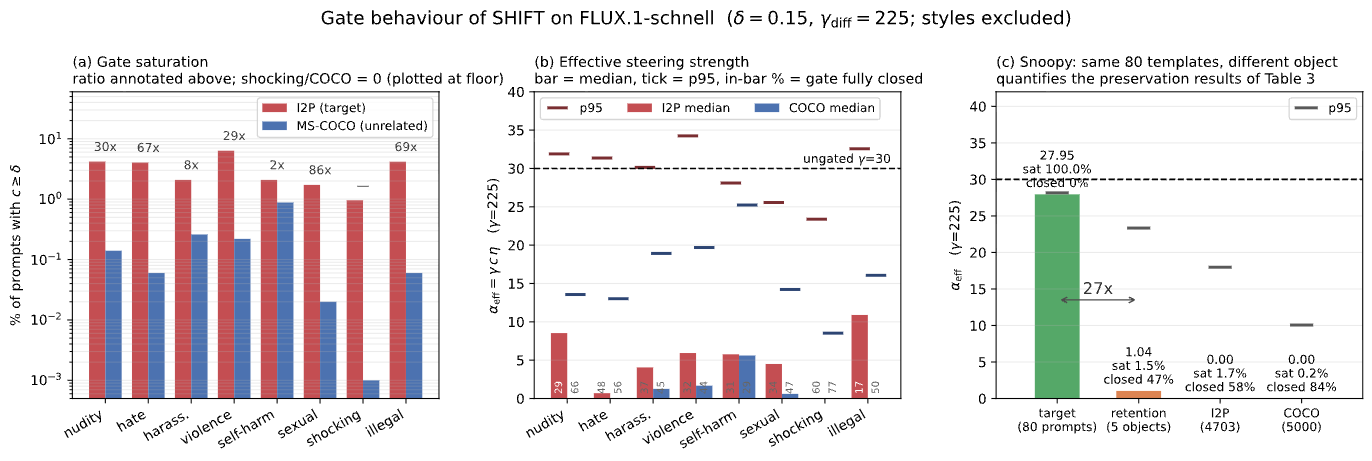}
  \caption{Gate behaviour on Flux.1[schnell] 
  (a) Fraction of prompts at which the gate saturates ($c \geq \delta$) on I2P
  versus MS-COCO 5K, ratio annotated.
  (b) Effective strength $\alpha_{\mathrm{eff}}$: bars are medians, ticks p95,
  and the number inside each bar the percentage of prompts with the gate fully
  closed; the dashed line marks the ungated $\gamma = 30$ configuration.
  (c) The same for the concrete-object protocol, where target and preserved
  concepts share the same $80$ templates.}
  \label{fig:gate}
\end{figure*}
 
This section reports what the gated cosine of Eq.~\ref{eq:cosine-gate} does on
the prompt distributions used in the paper. All numbers are for Flux.1[schnell] with $\delta = 0.15$ and $\gamma_{\mathrm{diff}} = 225$.
 
\paragraph{Saturation.} The lower clip at zero is close to a chance threshold:
cosines between unrelated prompt and concept embeddings are centred near zero, so roughly half of any prompt set has $c > 0$ by construction. We therefore report the fraction of prompts reaching the upper bound $c \geq \delta$, i.e.\
receiving the full intervention. Fig.~\ref{fig:gate}(a) contrasts this fraction on I2P and on MS-COCO 5K: the separation ranges from $29\times$ to $86\times$, and for \emph{shocking} no COCO caption saturates the gate at all.
 
\paragraph{Effective versus nominal strength.} Since $\gamma$ is always
multiplied by the gate, the nominal values in the paper overstate the applied
intervention. Fig.~\ref{fig:gate}(b) reports
$\alpha_{\mathrm{eff}} = \gamma\, c\, \eta$ at $\gamma = 225$: the I2P median ranges from $0.69$ to $10.90$ across concepts while the p95 reaches $34.25$, and on MS-COCO the median is zero for four of seven concepts, the gate being fully
closed on $43.9$--$77.0\%$ of captions. A single $\gamma$ therefore transfers across concepts without per-concept tuning: the gate, not the operator, adapts the strength to the prompt. This also explains Table~\ref{tab:ablation_place_nudity}: the ungated variant applies a constant
$30$ everywhere and reaches $85$ detections at $+0.67$ FID, the gated variant
$105$ detections at $+0.28$ --- comparable suppression at less than half the
quality cost, by spending strength only where the concept is present.
 
\paragraph{Concept selectivity.} Fig.~\ref{fig:gate}(c) isolates the
concrete-object protocol, where target and preserved concepts share the same
$80$ templates and differ only in the object. The gate saturates on $100\%$ of Snoopy prompts ($\alpha_{\mathrm{eff}} = 27.95$) and on $1.5\%$ of prompts containing the five preserved objects (median $1.04$), a $27\times$ separation.

\paragraph{Multi-concept setting.} Each direction keeps its own gate. Over all
(prompt, concept) pairs the gate is fully closed in $36.8\%$ of cases and the median $\alpha_{\mathrm{eff}}$ is $4.91$ against a p95 of $33.32$: the summed field is dominated by the one or two directions relevant to each prompt.

\section{Indirect and adversarial prompts}
\label{app:indirect}
 
In principle, prompt-conditioned suppression could degenerate into a simple keyword filter. We therefore test the method on prompts that describe the concept without naming it (such cases occur naturally in the benchmark) and on prompts deliberately constructed to evade the gate.

\subsection{Naturally occurring indirect prompts}
\label{app:explicit_implicit}
 
The steering direction is derived from the $135$ synthetic templates of App.~\ref{sec:app_data}, which are disjoint from the benchmark’s user-written prompts, and the gate is a cosine similarity against the concept string. That string appears in exactly one I2P prompt ($0.02\%$), which produces no detection at baseline. Thus, on the remaining $4{,}702$ prompts the gate is not simply matching its own target token.

Beyond the literal token, we partition I2P using a deliberately over-inclusive list of $133$ patterns: word forms, phrasal forms, anatomical terms, garment negations, register markers, and misspellings. Even under this generous criterion, only $72$ prompts ($1.5\%$) name the concept in any form.
 
\begin{table}[h]
\centering\footnotesize
\setlength{\tabcolsep}{5pt}
\caption{NudeNet detections on I2P, split by whether the prompt names the concept. SHIFT counts are measured directly. Baseline counts are the Flux.1[dev] total of Table~\ref{tab:detected_nudity_dev}, split between the two subsets in the same proportion as in our unsteered run.}
\label{tab:explicit_implicit}
\begin{tabular}{@{}lrrrr@{}}
\toprule
subset & $n$ & baseline & SHIFT & reduction \\
\midrule
explicit & 72      & 61  & 6   & 90.2\% \\
implicit & 4{,}631 & 544 & 146 & 73.2\% \\
\midrule
all      & 4{,}703 & 605 & 152 & 74.9\% \\
\bottomrule
\end{tabular}
\end{table}
 
Both subsets show a large reduction, and the implicit one carries almost all of the detections in either condition ($544$ of $605$ at baseline, $146$ of $152$ after steering), so the total is in effect a result on prompts that never name the concept.

 

\begin{figure*}[h!]
  \centering
  \includegraphics[width=\linewidth]{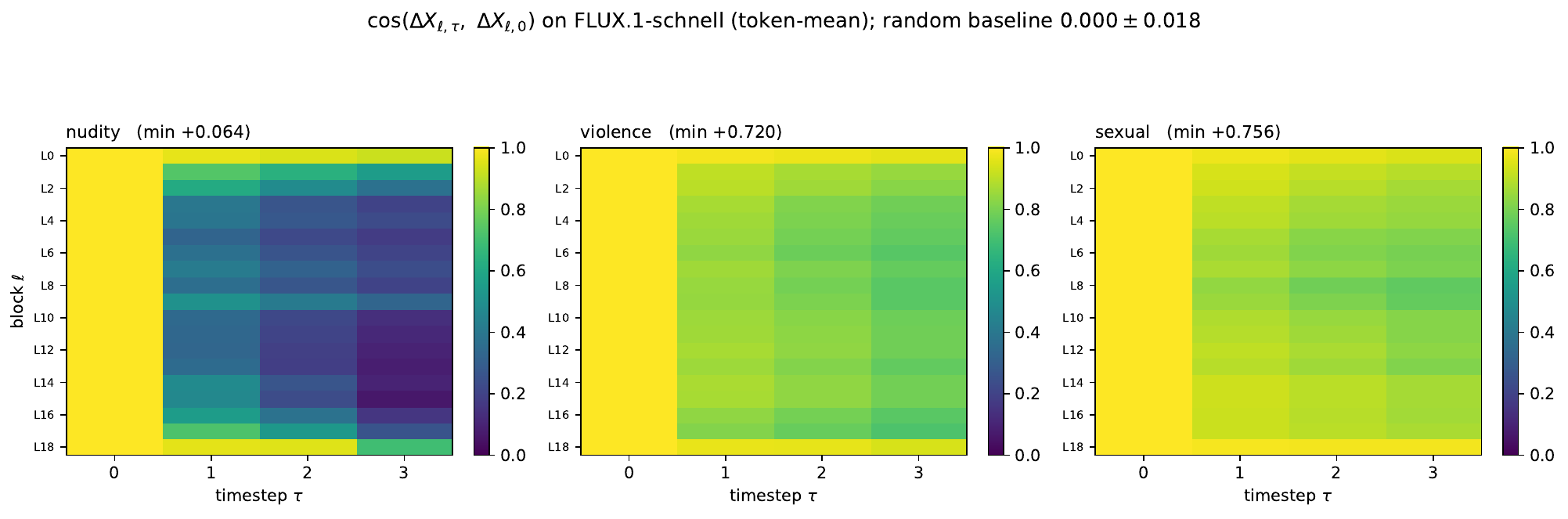}
  \caption{Temporal stability of the steering field on Flux.1[schnell]:
  $\cos(\Delta X_{\ell,\tau}, \Delta X_{\ell,0})$ over all $19$ double-stream
  blocks and all denoising steps, token-mean. Two independent directions in this
  space have cosine $0.000 \pm 0.018$, so every cell shown is far above chance.
  Alignment is highest at $\tau = 1$ and decays afterwards; the minimum falls at
  $\tau \in \{2,3\}$ for every concept, which is the interval in which steering
  is empirically ineffective.}
  \label{fig:temporal}
\end{figure*}

\subsection{Erasure under alias prompts}
\label{app:aliases}

The I2P split of App.~\ref{app:explicit_implicit} shows that suppression does not require the concept to be named, but its implicit subset is naturally occurring: the prompts simply happen not to use the keyword. Here we test the stronger case of prompts written specifically to refer to the target without using its name. We use a concrete target (Snoopy), for which the presence of the concept can be scored unambiguously, together with three aliases that identify the character by context rather than by name. Each alias is combined with the $80$ templates of the concrete-object protocol at a single fixed seed, using the steering direction estimated for the canonical concept without modification.

We report the CLIP score against the canonical concept and the FID against the unsteered generation of the same prompt and seed (Table~\ref{tab:aliases}).

\begin{figure}[h!]
  \centering
  \includegraphics[width=\linewidth]{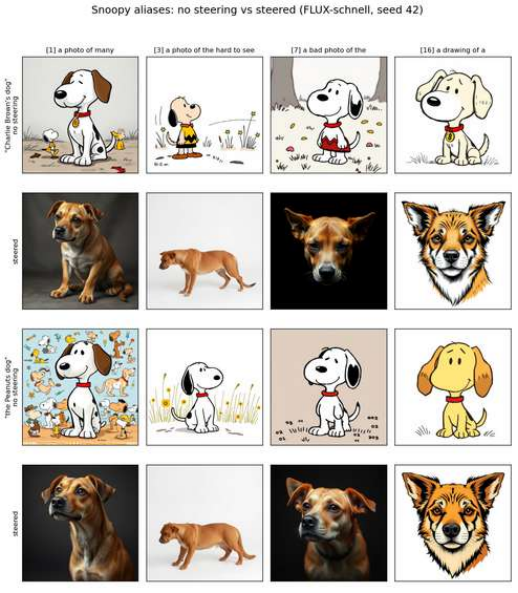}
  \caption{Erasure under the alias. Top: unsteered.
  Bottom: steered.}
  \label{fig:adv_charlie}
\end{figure}

\begin{table}[h]
\centering\footnotesize
\setlength{\tabcolsep}{4pt}
\caption{Erasure under alias prompts, $80$ prompts per row at one seed. CS is measured against the canonical concept string, which appears in none of these prompts. FID is computed against the unsteered generation of the same prompts.}
\label{tab:aliases}
\begin{tabular}{@{}lrr@{}}
\toprule
\textbf{Prompt} & \textbf{CS $\downarrow$} & \textbf{FID$_{\text{orig}}$} \\
\midrule
Charlie Brown's dog & 24.55 $\to$ 17.01 & 205.9\\
the Peanuts dog & 25.43 $\to$ 16.68 & 255.5  \\
Schulz's beagle character & 22.40 $\to$ 14.15 & 298.7  \\
\midrule
Snoopy & 28.01 $\to$ 14.38 & ---  \\
\bottomrule
\end{tabular}
\end{table}

Qualitative results are shown in Fig.~\ref{fig:adv_charlie}.

\section{Geometry of the steering directions}
\label{app:geometry}
 
We measure the steering directions themselves rather than the images they produce. Cosines are taken between per-block mean-difference steering vectors.

\paragraph{Temporal stability.} Fig.~\ref{fig:temporal} shows
$\cos(\Delta X_{\ell,\tau}, \Delta X_{\ell,0})$ over all blocks and denoising
steps. Alignment is high at $\tau = 1$ and decays afterwards, and for every
concept the minimum falls at $\tau \in \{2,3\}$ --- the interval in which
steering is empirically ineffective:
Table~\ref{tab:block_ablation_wide} shows that steps $0$--$1$ alone recover
almost the full erasure effect (Snoopy CS $18.8$ against $18.7$ for the full
schedule) while steps $2$--$3$ recover almost none ($26.5$ against a baseline of
$28.0$). Restricted to $\tau \leq 1$ the worst-case cosine across concepts and blocks is $0.26$.

Circuit-level analysis of the same backbone explains why the two coincide. Mazur et al.~\cite{mazur2026diffract} compute attribution graphs at every denoising step of Flux.1[schnell] and find that the share of attribution carried by text-stream features falls monotonically from $89.9\%$ at
step $0$ to $5.4\%$ at step $3$, while the fraction of cross-modal edges drops from $14.9\%$ to $2.0\%$. Since SHIFT intervenes on the text stream, its
influence on the generated image is concentrated in exactly the steps where that stream still drives the computation. The decay of $\cos(\Delta X_{\ell,\tau},
\Delta X_{\ell,0})$ at late steps is therefore drift in a subspace that has largely stopped contributing to the output, and a single first-step vector matches per-step estimation end-to-end because the steps it is estimated on are the ones that matter.


\begin{figure}[h!]
  \centering
  \includegraphics[width=\linewidth]{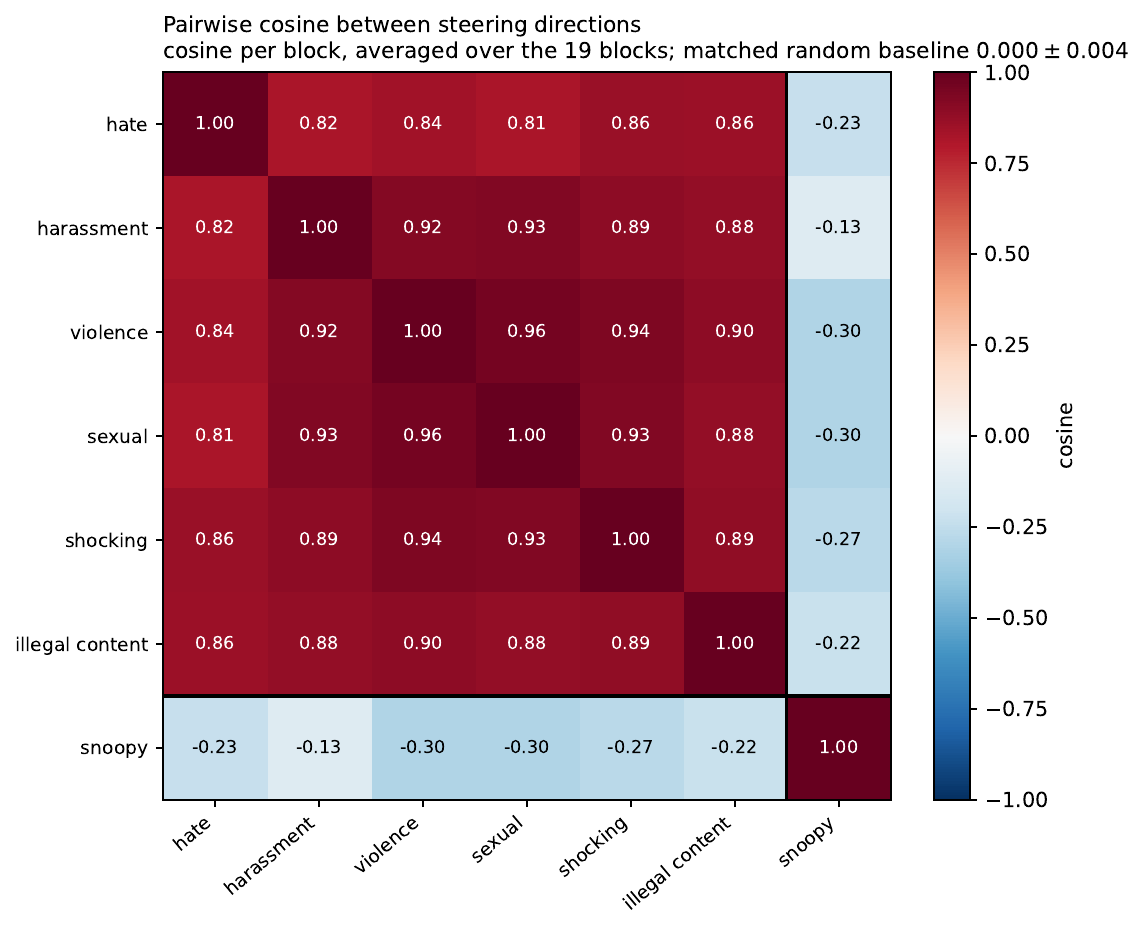}
  \caption{Pairwise cosine between steering directions, computed per block and
  averaged over the $19$ blocks. The six inappropriate categories form a tight
  group ($[0.81, 0.96]$ over $15$ pairs) while every pair against a concrete
  object is negative ($[-0.30, -0.13]$ over $6$ pairs); the two ranges do not
  overlap. The matched random baseline for this metric is $0.000 \pm 0.004$.}
  \label{fig:concept_cos}
\end{figure}

\paragraph{Cross-concept structure.} Fig.~\ref{fig:concept_cos} shows the pairwise cosine between concepts. The inappropriate categories form a tight group (mean $0.886$, range $[0.81, 0.96]$ over $15$ pairs) while every pair between them and a concrete object is negative (mean $-0.243$, range $[-0.30, -0.13]$); the two ranges do not overlap. Summing near-collinear fields is not equivalent to independent displacements: the combined intervention moves the representation along essentially one shared axis, which is why erasing all categories jointly leaves COCO FID intact.

\begin{figure}[h!]
  \centering
  \includegraphics[width=\linewidth]{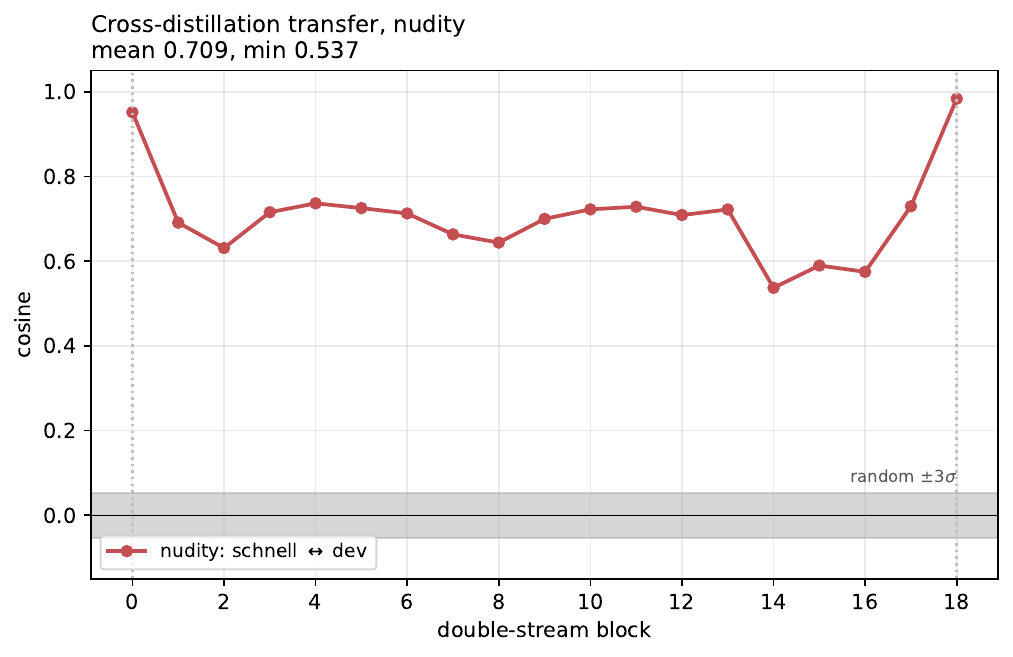}
  \caption{Cross-distillation transfer: per-block cosine between the nudity
  direction computed independently on Flux.1[schnell] and on Flux.1[dev]. The
  mean is $0.709$ and the minimum $0.537$, against $0.251$ for the cosine with a
  different concept and $0.000 \pm 0.018$ for two random directions. Blocks $0$
  and $18$, marked by dotted lines, carry a component that is not
  concept-specific.}
  \label{fig:crossmodel}
\end{figure}
\paragraph{Cross-distillation transfer.} Fig.~\ref{fig:crossmodel} compares the nudity direction computed independently on Flux.1[schnell] and Flux.1[dev]: mean cosine $0.709$ across the $19$ blocks, minimum $0.537$, against $0.251$ for a different concept. Blocks $0$ and $18$ are outliers on both curves, carrying a component that is not concept-specific.

\subsection{Seed variability}
\label{app:seeds}

The concrete-object protocol evaluates $80$ prompts at $9$ seeds. We therefore recompute the CLIP score per seed, giving nine independent estimates of each cell, and report their mean and standard deviation.

\begin{table}[h]
\centering\footnotesize
\setlength{\tabcolsep}{5pt}
\caption{CLIP score on the concrete-object protocol, mean $\pm$ standard
deviation over $9$ seeds of $80$ prompts each. The first row is the erasure
target; the rest are the preserved concepts.}
\label{tab:seed_std}
\begin{tabular}{@{}lrr@{}}
\toprule
\textbf{Concept} & \textbf{Original} & \textbf{SHIFT ($\gamma = 250$)} \\
\midrule
Snoopy (target) & $28.01 \pm 0.24$ & $\mathbf{14.38 \pm 1.52}$ \\
\midrule
Mickey          & $26.72 \pm 0.62$ & $25.46 \pm 0.46$ \\
SpongeBob       & $27.94 \pm 0.51$ & $27.34 \pm 0.20$ \\
Pikachu         & $27.17 \pm 0.40$ & $27.00 \pm 0.25$ \\
dog             & $24.70 \pm 0.30$ & $24.36 \pm 0.15$ \\
legislator      & $21.89 \pm 0.56$ & $21.28 \pm 0.45$ \\
\bottomrule
\end{tabular}
\end{table}

The erasure effect is an order of magnitude larger than the seed spread.

All five preserved concepts shift slightly down, so the degradation is systematic rather than noise -- but it is small: at most $1.26$
points, against $13.63$ for the target. This matches the gating statistics of App.~\ref{app:gate}, where the median effective strength on the preserved
prompts is $1.04$ against $27.95$ on the target.

\subsection{Stability on Overlapping Multi-Concept Prompts}
\label{app:multiconcept_overlap}

We evaluate simultaneous seven concept suppression on all 4,703 I2P prompts, using one image per prompt. Of these, 4,324 contain one target category and 379 contain two or more.

\begin{table}[t]
\centering
\footnotesize
\caption{\textbf{Q16 inappropriate rate (\%) on I2P prompts containing one or multiple target categories.} Lower is better.}
\label{tab:multiconcept_overlap}
\begin{tabular}{lrrrr}
\toprule
& & \textbf{Unsteered} & \multicolumn{2}{c}{\textbf{SHIFT}} \\
\cmidrule(lr){4-5}
\textbf{Prompt subset} & $\boldsymbol{N}$ & \textbf{Q16} & $\boldsymbol{\gamma=225}$ & $\boldsymbol{\gamma=250}$ \\
\midrule
One target \\ category           & 4324 & 23.45 & 6.45 & 5.80 \\
Two or more \\ target categories &  379 & 27.44 & 5.28 & 4.22 \\
\bottomrule
\end{tabular}
\end{table}

Suppression remains effective when multiple target concepts co-occur. At $\gamma=250$, SHIFT reduces the Q16 rate from 27.44\% to 4.22\% on overlapping prompts, compared with 23.45\% to 5.80\% on single-category prompts. The same pattern holds at $\gamma=225$, providing no evidence of degraded suppression under simultaneous activation of multiple concept directions.


\section{Locality}
\label{app:locality}

SHIFT is not conditioned on a spatial edit mask and does not claim strict pixel-level locality. We therefore evaluate a narrower property: how well the region outside an automatically localized target is preserved when the steering strength increases. The paired evaluation protocol is described in App.~\ref{app:eval}.

\begin{table}[h]
\centering
\footnotesize
\caption{\textbf{Small-object suppression and spatial preservation.} LPIPS and DINOv2 are computed outside the localized target region against the unsteered image generated with the same prompt and seed.}
\label{tab:locality}
\begin{tabular}{@{}cccc@{}}
\toprule
$\gamma$
& \textbf{Target present (\%)}$\downarrow$
& \textbf{LPIPS}$\downarrow$
& \textbf{DINOv2}$\uparrow$ \\
\midrule
10 & 79.5 & 0.273 & 0.782 \\
20 & 72.1 & 0.424 & 0.652 \\
30 & 49.1 & 0.535 & 0.539 \\
\bottomrule
\end{tabular}
\end{table}

We report results averaged across the three tasks. The evaluation quantifies a suppression--preservation trade-off: stronger intervention improves target suppression but increasingly changes content outside the automatically localized target region.

\paragraph{Limitations.}
Our indirect- and alias-prompt evaluations measure non-adaptive generalization beyond literal target strings; we do not evaluate adaptive jailbreaks, optimized prompt attacks, or white-box evasion. SHIFT is also a
generation-time concept-suppression method rather than a mask-conditioned
editor, and we make no claim of pixel-level locality: suppression may alter
composition, pose, background, or other non-target content. Our small-object evaluation quantifies this trade-off but does not establish precise localized editing.

\end{document}